\documentclass[11pt]{article}
\usepackage[margin=1in]{geometry}
\usepackage{amsmath,amssymb,amsthm}
\usepackage{booktabs}
\usepackage{graphicx}
\usepackage{xcolor}
\usepackage{natbib}
\usepackage[colorlinks=true,linkcolor=blue,citecolor=blue,urlcolor=blue]{hyperref}

\newtheorem{theorem}{Theorem}
\newtheorem{lemma}{Lemma}
\newtheorem{proposition}{Proposition}
\newtheorem{corollary}{Corollary}
\theoremstyle{definition}
\newtheorem{definition}{Definition}
\newtheorem{assumption}{Assumption}
\theoremstyle{remark}
\newtheorem{remark}{Remark}

\newcommand{\paperfig}[4]{\begin{figure}[t]\centering\IfFileExists{figures/#1.pdf}{\includegraphics[width=#2\linewidth]{figures/#1.pdf}}{\fbox{\parbox[c][3cm][c]{0.85\linewidth}{\centering\small\texttt{\detokenize{figures/#1.pdf}} (run make\_figures.py)}}}\caption{#3}\label{#4}\end{figure}}

\title{A Conditional No-Hair Theorem for Transformers:\\ How Language Models Forget Their Prompt}
\author{Shuyang Xiang\\ Independent researcher}
\date{October 2026}

\begin{document}
\maketitle

\begin{abstract}
A black hole retains only a few conserved charges of the matter it formed from. We are inspired to make the analogy and we ask how much an autoregressive transformer retains of its prompt after generating $t$ tokens. For two prompts followed by the same continuation, the difference of their hidden states decays approximately as a power law $t^{-\alpha}$, where $\alpha=1$ is the rate at which a fixed prompt's share of uniform attention shrinks. Beyond the first layers, trained models forget faster ($\alpha\approx0.55$--$0.7$) than untrained ones of the same architecture ($\alpha\approx0.2$--$0.5$). For finite-depth pre-LN softmax transformers with bounded attention logits and non-degenerate pre-norm activations, we prove that this offset is $O\big(P(\log t)^{L-1}/t\big)$ for prompt length $P$ and depth $L$: the prompt's direct influence is diluted as the context grows, and \emph{relay} through continuation tokens adds only logarithmic factors. Conditional on the generated text, the prompt's influence on the next token thus vanishes. The bound fixes only the asymptotic shape and is uninformative at the lengths we measure in trained networks; an exactly solvable linear relay model shows that it cannot be improved and accounts for the untrained curves. Cancelling the dilution with a logit bias on the prompt stops the decay early in training, and training suppresses relay by about one billion tokens, before attention sinks form. A toolkit that measures $\alpha$ for any Hugging Face causal language model is publicly available.
\end{abstract}
\section{Introduction}
\label{sec:intro}

Think of a conversation that runs long with your friends. You open with a remark about the weather: say, \emph{Wednesday looks nice}, or \emph{Thursday looks nice}. By the time it has moved through work, a film, a friend's illness, nobody could tell from the last ten minutes which of the two weekdays you said. The opening line is still in the room, in the sense that nothing was deleted. It has simply stopped mattering. Large language models behave the same way. A system prompt sits at the start of the context, all of its tokens visible to every later query, and yet over a few dialogue turns the persona it sets drifts measurably \citep{li2024instability}. An agent whose constraints are given once, at the start of a long trajectory, undergoes the same erosion. Nothing is truncated, and still the model somehow stops listening. We find that the offset between the hidden states of two prompts, measured under a shared continuation, decays close to a power law, and that, counterintuitively, training accelerates this decay: in the layers beyond the first, a trained network forgets its prompt \emph{faster} than a random-weight network of the same architecture. Whatever information about the prompt persists in the generated text, such as its language, is then plausibly carried by the text itself rather than by a memory in the hidden state; Section~\ref{sec:exterior} tests this.

Why should an opening remark fade at all, when it never leaves the context? Physics offers a template. A black hole forgets almost everything about the matter that formed it: an exterior observer can only measure a few charges, and these persist because the exterior field itself can carry them. The corresponding questions for a transformer are then clear.

The core inquiry of this paper is simple to state. Give a language model a prompt, let it write $t$ tokens, and compare two prompts that differ in a single word, in phrasing, in language, or in genre. Long after the prompt, can anything inside the model still tell them apart? And if not, why not?

We call the residual amount of the prompt in the hidden state its \emph{imprint}, and ask three questions about it: how large is the imprint after $t$ generated tokens, how fast does it decay, and what properties of the prompt can still be read off from it. Whether an imprint of a given size is readable depends on how it compares with the variation that the continuation itself causes, so we also track a \emph{decodability} ratio: the imprint measured against the spread of hidden states across different continuations. The two quantities are not the same, but they are linked --- a decoder can read the prompt off the state only while the imprint stands out against the continuation --- and we show in what follows that in practice they decay at the same rate, which lets a single exponent summarise how quickly a prompt is forgotten.

This paper answers these questions in three steps. We first define a quantity that can be measured: the \emph{prompt offset}, the difference between the hidden states of two runs that share a continuation but differ in their prompt. We then prove that, under two conditions on the architecture (bounded attention logits and pre-norm activations bounded away from zero), this offset is bounded by a quantity that decays as $t^{-1}$ up to logarithmic factors, and show that the bound is attained by a model that can be solved exactly. Finally, we trace where the prompt's influence goes as it decays: through the attention path we call \emph{relay}, which training suppresses by about one billion tokens.

\paragraph{How the theorem and the measurements relate.} The theorem fixes only the asymptotic shape of the decay. Its constant is so large for trained networks that the bound is uninformative within any window we measure (Remark~\ref{rem:constant}), so the measured exponents of trained models, $\alpha\approx0.55$--$0.7$, are not in conflict with the asymptotic rate $1$: the bound constrains the decay only for $t\gg Pe^{\Delta}$, far beyond our windows. These exponents are measurements, not predictions of the theorem; the solvable model of Section~\ref{sec:theory-linear} accounts for the untrained networks.

\paragraph{Contributions.} The imprint is measured as a prompt offset, and five results characterise how it decays.

\begin{enumerate}
  \item \textbf{The measured decay.} We define the teacher-forced prompt offset and show that it decays approximately as $t^{-\alpha}$, with logarithmic corrections in untrained networks and close to a pure power law in trained Pythia on coherent text. The spread across continuations stays essentially constant ($\beta\approx 0$ except in untrained Qwen), so the rate at which the prompt is forgotten is governed by a single exponent (Section~\ref{sec:measure}).
  \item \textbf{A theoretical bound.} For finite-depth pre-LN transformers with bounded attention logits and non-degenerate pre-norm activations (Assumption~\ref{ass:nondeg}, verified empirically on the sequences we measure), the offset is $O\big(P(\log t)^{L-1}/t\big)$; conditional on the generated text, the prompt's influence on the next token vanishes at the same rate. For trained networks the constant makes the bound uninformative within our measurement windows. Uniform attention attains the bound asymptotically, and the resulting linear relay model is solved exactly for arbitrary layer gains (Section~\ref{sec:theory}).
  \item \textbf{The relay mechanism.} We decompose the attention contribution to the offset into a direct path (queries reading the prompt) and a relay path (queries reading continuation tokens that already carry prompt information). The decomposition reproduces the measured $\alpha$ within $0.06$ at every layer and training stage (Section~\ref{sec:mechanism}).
  \item \textbf{From mechanism to training.} Compensating the dilution with a logit bias on the prompt removes or reverses forgetting; on Pythia-160m, forgetting sets in between steps 128 and 512 (roughly 0.3--1B training tokens), before attention sinks appear; attention weights alone are a misleading proxy (Sections~\ref{sec:mech-training} and~\ref{sec:causal}).
  \item \textbf{An application, with its costs.} A mild $\log t$ compensation increases instruction adherence at a small perplexity cost but induces instruction echoing; in multi-turn conversations it shows a consistent trend towards higher adherence, but no more than a constant boost with the same mean bias, so the benefit comes from extra attention to the instruction rather than from the shape of the compensation (Section~\ref{sec:application}).
\end{enumerate}
\section{Related work}
\label{sec:related}

\paragraph{Where this paper sits.} Three lines of work are close to ours. One studies the practical failure of long contexts: instruction drift, attention sinks, and the limits of what a model can generate. The second studies information flow through attention, both empirically and as over-squashing. The third studies the statistical structure of long text and the power-law decay of attention. Our question ``how much of a prompt remains in the hidden state after $t$ generated tokens?''  sits between these lines, and our answer borrows from each: the failure mode is from the first, the path decomposition from the second, the power law from the third.

\paragraph{Instruction drift.} \citet{li2024instability} show that adherence to a system prompt degrades within a few dialogue rounds, attribute it to attention decay, and propose split-softmax. Our causal experiment agrees on the cause, and our quantitative law explains why the drift occurs: the prompt's share of attention shrinks as $1/t$ under uniform attention, so the decay is a geometric consequence of a growing context rather than an artefact of any particular dialogue pattern.

\paragraph{Attention sinks and over-mixing.} Sinks absorb attention without carrying information \citep{xiao2024streaming,gu2025sink}; \citet{barbero2025first} argue that they prevent over-mixing. Our relay path is a form of mixing, and our training-dynamics measurement indicates the ordering in Pythia-160m: forgetting sets in between steps 128 and 512 (roughly 0.3--1B tokens), while the attention mass on the first token stays at or below $0.003$ through step 1000 (about 2B tokens), so sinks are not the initial cause of forgetting in this model.

\paragraph{Information flow.} Norm-based attention analysis \citep{kobayashi2020norm} and attention flow \citep{abnar2020flow} motivate our path decomposition; rank collapse \citep{dong2021rank} is a depth-wise counterpart of the token-wise mixing studied here. \citet{barbero2024glasses} show that in decoder-only transformers the sensitivity of the last token to earlier tokens is limited by over-squashing; our theorem makes this loss quantitative for a fixed prompt as generation proceeds, and the linear relay model shows that the $t^{-1}$ rate cannot be improved in the uniform-attention regime.

\paragraph{Signal propagation and mean-field theory.} The propagation of a perturbation through depth in randomly initialised networks has been analysed with mean-field methods as a question about correlation lengths and critical initialisations \citep{poole2016exponential,schoenholz2017deep}; for transformers, \citet{noci2022signal} study signal propagation and rank collapse at initialisation. In our setting the perturbation is confined to a fixed set of positions (the prompt) and the control parameter is the context length, so the relevant decay is along the token axis, and the exponent $\alpha$ plays the role of a decay exponent along that axis. The untrained networks, whose weights carry no statistics of any corpus, are the natural counterpart of those mean-field analyses.

\paragraph{Long-range statistics of text.} Mutual information in natural text decays as a power law \citep{lin2017critical,chen2025l2m}. A power-law forgetting curve could in principle be inherited from the data. Our random-initialisation baseline rules this out for the existence of the power law: an untrained network, whose weights carry no statistics of any corpus, already forgets as a power law, with exponents that are unchanged (to within $0.01$) when the continuation sentences are shuffled. In trained networks the exponent does depend on the continuation text (Section~\ref{sec:results-struct}), but the ordering of the axes and the trained--untrained contrast do not.

\paragraph{Prompt influence on generated text.} \citet{prodperc2026} rescore a fixed generation under different prompts and find that the influence of the prompt fades with accumulated context. Where they measure the effect on the output distribution, we measure it on the hidden state, which lets us track its decay layer by layer and separate the direct path from relay.

\paragraph{Power-law decay of attention.} Attention scores to earlier tokens are known to decay roughly as a power law over generation steps, which \citet{siftattention2025} exploit for dynamic sparse attention. Our dilution lemma gives a mechanism for such decay; the path decomposition shows, however, that attention weights are an unreliable proxy for information flow, because sinks attract weight without carrying prompt-specific content, and the decay of the trace in the hidden state is what determines whether the prompt is still readable.

\paragraph{Limits of long generation.} \citet{accessible2026} bound which output sequences a transformer can generate at all, finding a sharp length threshold beyond which most sequences become inaccessible. Their question is what a model can produce; ours is the complementary one of how the influence of a given prompt on what it produces decays.

\paragraph{Attention fading and length-dependent scaling.} \citet{nakanishi2025ssmax} observes that softmax attention flattens as the context grows (``attention fading'') and proposes Scalable-Softmax, which multiplies the logits by $s\log n$. Our dilution lemma is a quantitative form of the same effect. A logit scaling that grows with context length makes the logits unbounded, so Lemma~\ref{lem:consts}(a) and with it Lemma~\ref{lem:dilution} no longer apply (see the scope paragraph in Appendix~\ref{app:proofs}); our analysis therefore predicts that such scalings, including Scalable-Softmax, retain prompt information longer.

\section{Setup and measurement}
\label{sec:measure}

We want to know whether a model that has generated $t$
tokens after a prompt still carries that prompt in its hidden state, and how fast this memory fades as $t$
 grows. Answering it requires three choices: what to compare, so that a difference in hidden states can only come from the prompt (Section~\ref{sec:measure-setting}); what to measure, so that the result says whether the prompt can be read off the state (Section~\ref{sec:measure-def}); and how to estimate decay rates reliably (Section~\ref{sec:measure-est}). 

\subsection{What to compare}\label{sec:measure-setting}

Let a decoder-only language model receive a prompt $x$ of $P$ tokens and then generate autoregressively, without further input. After $t$ generated tokens, information about $x$ can be held in two places. The first is the generated text, which any observer can read: call this the \emph{exterior}. The second is the residual stream at the current position, which also sees $x$ through attention: call this the \emph{interior}. Only the interior is a memory of the network; the exterior is a record anyone could consult. This distinction is the same one that separates a black hole from its exterior field, and it will return when we ask what survives.

The natural experiment, letting two prompts generate freely and comparing their hidden states, cannot tell these apart: two prompts produce two different texts, and a difference in hidden states may be a memory of the prompt or merely a response to different text. We therefore fix the text: both prompts are followed by the same continuation $c$, written independently of either, and we compare the hidden states along it. With the continuation held fixed, any remaining difference between the two runs must come from the prompt. This is also the quantity bounded by Theorem~\ref{thm:nohair}, and the one that must vanish for a no-hair statement to hold.

The measurements in this section and in Sections~\ref{sec:theory}--\ref{sec:application} therefore use what we call \emph{teacher-forcing}: the continuation is written by a human, not by the model. This is a measurement device, not the failure mode itself. The failure mode is the one that occurs in free generation, when the continuation is produced by the model and the prompt's influence on the next token is expected to travel mainly through the text it has caused (Corollary~\ref{cor:exterior} and Section~\ref{sec:exterior} examine what this does and does not imply). Section~\ref{sec:exterior} returns to that setting and asks what survives. In the theoretical sections that follow, we will replace the time index \(t\) with the distance \(i\) from the end of the prompt, so that \(\delta_l(i) \equiv \delta_l(t; c)\).

\subsection{What to measure}\label{sec:measure-def}

The most direct imprint of the prompt is the difference between the two hidden states. Fix a layer $l$, and let $h_l(x\circ c,t)\in\mathbb R^d$ be the residual stream after block $l$ (before normalisation) at time $t$ along $c$, where $d$ is the hidden dimension and $\circ$ denotes concatenation, so that $x\circ c$ is the prompt followed by the continuation. To reduce token-level fluctuations we average it over a fixed window of consecutive continuation tokens and assign it the time $t$ of the window's centre.

\begin{definition}[Prompt offset]\label{def:offset}
For two prompts $x,x'$ and a shared continuation $c$, the prompt offset at layer $l$ and time $t$ is
\[
\Delta_l(t;c)=h_l(x'\circ c,t)-h_l(x\circ c,t),\qquad\text{with norm }\delta_l(t;c)=\|\Delta_l(t;c)\|.
\]
We write $\Delta_{l,j}$ for the $j$-th coordinate of $\Delta_l$, and $h_{l,j}$ for the $j$-th coordinate of $h_l$.
\end{definition}

A small offset is not by itself evidence of forgetting: what matters is whether it stands out against the variation that the continuation itself causes. If the offset is the same for every continuation, the prompt can be read off the state whatever text follows; if it is swamped by the differences between continuations, it cannot. We therefore split the offset into a part common to all continuations, the \emph{signal}, and compare it with the spread of hidden states across continuations, the \emph{noise}. Both are measured in units of $\sigma_{l,j}$, the standard deviation of $h_{l,j}$ across continuations in the first window, pooled over both prompts, so that no coordinate dominates because of its scale. We write $n_c$ for the number of continuations per pair, and $\mathbb E_c$ and $\operatorname{Var}_c$ for the mean and variance across them.

\begin{definition}[Signal, noise and decay exponents]\label{def:signal}
The signal and the noise at layer $l$ are
\begin{equation}\label{eq:SN}
  \mathcal S_l(t)=\frac1d\sum_{j=1}^d\frac{\big(\mathbb E_c\,\Delta_{l,j}(t;c)\big)^2}{\sigma_{l,j}^2},\qquad
  \mathcal N_l(t)=\frac1d\sum_{j=1}^d\frac{\operatorname{Var}_c\,h_{l,j}(x\circ c,t)}{\sigma_{l,j}^2},
\end{equation}
and we summarise how fast each decays by a power-law exponent: $\mathcal S_l(t)\propto t^{-2\alpha_l}$ and $\mathcal N_l(t)\propto t^{-2\beta_l}$.
\end{definition}

The exponent we care about is $\alpha_l$: it measures how quickly the signal of the prompt fades. The exponent $\beta_l$ measures how quickly the noise fades as well, and we track it because a prompt is only decodable while the signal stands out against the noise: decodability scales as $t^{-2(\alpha_l-\beta_l)}$. We find $\beta_l\approx0$ almost everywhere (Section~\ref{sec:results-alpha}), so the rate at which the prompt is forgotten is the single number $\alpha_l$.

\subsection{Estimating decay rates}\label{sec:measure-est}

\paragraph{Sampling times.} Decay laws are best seen on a logarithmic time axis, so we space measurements by octaves. For Pythia on real-corpus pairs, windows end at $t_e=16,32,\dots,1024$; for Qwen, at $t_e=16,\dots,2048$. For the hand-written scenes, which use Pythia and whose continuations are at least $1920$ tokens long, the same octave spacing runs to $t_e=1920$, short of Pythia's $2048$-token context. Each window covers the $16$ tokens ending at these points. We label it by its centre, not by its end: end labelling biases the fitted exponent upwards when a window spans a whole octave, as it does in the first octaves. For offsets of the form $(P+s)^{-\alpha}$, the bias is up to $0.19$ locally and $0.07$ globally with end labels, against less than $0.03$ and $0.01$ with centre labels; on our data the global exponents drop by $0.05$--$0.10$ when going from end to centre labels.

\paragraph{Signal.} The signal is the part of the offset common to all continuations, so we estimate it by averaging $\Delta_{l,j}$ over continuations and squaring. At late times this average is small, and the naive estimate $(\mathbb E_c\Delta_{l,j})^2$ is then dominated by its sampling bias $\operatorname{Var}_c\Delta_{l,j}/n_c$, which would make the decay look slower than it is. We therefore use the unbiased estimator
\[
\widehat{\mathcal S}_l(t)=\frac{1}{d\,n_c(n_c-1)}\sum_{a\neq b}\sum_{j=1}^d\frac{\Delta_{l,j}(t;c_a)\,\Delta_{l,j}(t;c_b)}{\sigma_{l,j}^2},
\]
which corrects for this bias by averaging over pairs of continuations rather than squaring the mean. We also report the exponent of the mean squared offset $\mathcal M_l(t)=\frac1d\sum_j\mathbb E_c\Delta_{l,j}^2/\sigma_{l,j}^2$. Unlike $\mathcal S_l$, which averages the offset before squaring and therefore cancels offsets whose direction varies across continuations, $\mathcal M_l$ averages the squared offset, so it captures both the common part and the variation in direction. Comparing the two tells us how much of the decay is decorrelation rather than shrinking norm (Remark~\ref{rem:jensen}).

We use $n_c=8$ continuations per pair for real-corpus pairs and $n_c=16$ per scene for hand-written scenes. All real-corpus exponents of Section~\ref{sec:results-alpha} use $\widehat{\mathcal S}$, and so do the path decomposition (Section~\ref{sec:mechanism}), the compensation experiment (Section~\ref{sec:causal}) and the trade-off (Section~\ref{sec:application}). The hand-written scenes use the plug-in estimate, because the correction requires storing pairwise products of offsets, which the earlier runs did not record; their late-time exponents are correspondingly biased low.
\paragraph{Exponents.} The signal is first averaged over prompt pairs (and, where stated, over axes); the global exponent $\alpha_l$ is the least-squares slope of $-\tfrac12\ln\widehat{\mathcal S}_l$ against $\ln t$ over all measurement points, with $95\%$ bootstrap intervals over scenes or prompt pairs. A single exponent can hide a decay that is not a pure power law, so we also compute the \emph{local exponent} $\alpha_{\mathrm{loc}}$, the same slope between consecutive points. A pure power law has constant $\alpha_{\mathrm{loc}}$; dilution with logarithmic corrections gives an $\alpha_{\mathrm{loc}}$ that rises towards $1$ (Section~\ref{sec:theory}). $\beta_l$ is estimated in the same way from $\mathcal N_l$.

\subsection{Prompts and continuations}\label{sec:measure-data}

Our picture predicts that the decay exponent $\alpha$ depends on the architecture and on the length of the prompt, but not on how different the two prompts are: a larger difference should raise the amplitude of the offset, not change how fast it decays. We test this prediction by varying the kind and size of the difference, on two sets of prompts.

\paragraph{Hand-written scenes.} Twenty scenes, each with a base prompt and four variants, from the smallest to the largest change: one content word changed (\emph{detail}), a paraphrase (\emph{para}), a news-wire rewrite (\emph{genre}) and a translation (\emph{lang}). Continuations are human-written: sixteen passages per scene from the WikiText-103 validation set, at least $1920$ tokens long, shared by all five prompts of the scene. The full list is in Appendix~\ref{app:prompts}.

\paragraph{Real-corpus pairs.} Hand-written prompts may share stylistic regularities, so we repeat the measurement on pairs drawn from public corpora, one hundred per axis: a WikiText sentence and the same sentence with one word replaced (\emph{detail}), MRPC paraphrase pairs (\emph{para}), WMT14 English--French pairs (\emph{lang}), and two unrelated WikiText sentences (\emph{unrelated}), the largest possible difference. Continuations are passages from the WikiText-103 validation set, eight per pair, shared by both prompts of the pair.

\paragraph{Shared first token.} Trained models often concentrate attention on the first token, an attention sink \citep{xiao2024streaming}. When two prompts begin with different tokens, the sink itself differs, and its contribution need not dilute. Where stated, we prepend a common token (\texttt{<|endoftext|>}) to both prompts to remove this difference.

\subsection{Models}\label{sec:measure-models}

Two questions guide the choice of models: when during training does the forgetting rate change, and does the picture depend on architectural details? Pythia \citep{biderman2023pythia} answers the first, since it provides training checkpoints of a fixed architecture (LayerNorm, parallel residual): we use Pythia-160m at eight checkpoints from initialisation to the end of training (steps $0$, $128$, $256$, $512$, $1000$, $4000$, $16000$, $143000$), and Pythia-410m and Pythia-1b at initialisation and after training, to separate the effects of training, width and depth. Qwen2.5 \citep{qwen2025} addresses the second with a different architecture (RMSNorm, sequential residual, grouped-query attention): we use Qwen2.5-0.5B, trained and at random initialisation (the library's default scheme: normal weights with standard deviation $0.02$, seed $0$; for Pythia, step $0$ is the released initial checkpoint), and the trained Qwen2.5-1.5B. All computations run in float32 on a single Apple-silicon laptop.

\section{A conditional no-hair theorem}
\label{sec:theory}

\subsection{Setting and notation}
\label{sec:theory-setting}

\paragraph{Architecture.} We work with $L$-layer pre-LN decoders: causal attention, residual connections, and rotary or no positional encoding inside the logits (as in Pythia and Qwen). The pre-LN placement matters here: each block normalises its input inside the residual branch, before attention and before the MLP, so the residual stream itself is never normalised. This is the sequential form, $h_l=h_{l-1}+\mathrm{Attn}(\mathrm{LN}(h_{l-1}))$ followed by the MLP update, as in Qwen, or the parallel form, where both updates are computed from $h_{l-1}$ and added, as in Pythia. The scope covers most current decoders, with LayerNorm or RMSNorm (see below); it excludes post-LN blocks, linear attention and recurrent state.

\paragraph{Notation.} We use the same setup as Section~\ref{sec:measure}, with one change of notation for the proof. Two prompts $x,x'$ are followed by the same continuation $c$, and $h_l(i)$, $h'_l(i)$ denote the residual streams at the $i$-th continuation token after $l$ blocks. This is the quantity $h_l(x\circ c,t)$ of Definition~\ref{def:offset}, with the time index $t$ replaced by the distance $i$ from the end of the prompt. Write $\delta_l(i)=\|h_l(i)-h'_l(i)\|$. Token embeddings of the continuation coincide, so $\delta_0(i)=0$.

The two prompts have lengths $P$ and $P'$. From here on, $P$ denotes the larger of the two, so that any bound in terms of $P$ applies to both. We index continuation positions by their distance $i\ge1$ from the end of the prompt, so the $i$-th continuation query attends to at most $P+i$ keys. With rotary encodings, positions enter only the logits, through relative offsets, so $P\neq P'$ is allowed; with absolute position embeddings we take $P=P'$.

Finally, we write the normalisation as $\mathrm{LN}(h)=\sqrt d\,\Pi h/\|\Pi h\|$. Here $\mathbf 1$ is the all-ones vector, $I$ the identity, and $\Pi=I-\tfrac1d\mathbf 1\,\mathbf 1^\top$ the projection onto the subspace orthogonal to it, so that $\Pi h=h-\bar h\,\mathbf 1$ with $\bar h$ the mean of $h$. Thus $\mathrm{LN}$ first removes the mean and then rescales to length $\sqrt d$. The gain and bias of the layer are absorbed into the adjacent linear maps and do not appear separately. RMSNorm is the special case $\Pi=I$, in which the mean is not removed.

The proof needs the normalisation layers to be well conditioned:

\begin{assumption}[Non-degenerate pre-norm activations]\label{ass:nondeg}
There is $r>0$ such that $\|\Pi h\|\ge r$ at every position and in both runs, whenever $h$ is the input of blocks $2,\dots,L$, the input of the final norm, or, in a sequential block, the post-attention residual that the MLP reads.
\end{assumption}

The assumption is what gives the normalisation Lipschitz: the constant in Lemma~\ref{lem:consts} is $2\sqrt d/r$, and it degenerates if $\Pi h$ can vanish. In a pre-LN residual stream it cannot do so in a harmful way. The stream is never normalised, and each block adds its update to it, so the norm grows with depth rather than collapsing; the only vectors that come close to the degenerate case are the token embeddings, which are excluded from the assumption. On the sequences we measured (up to $1024$ continuation tokens), the assumption holds at block inputs with a comfortable margin: $r/\sqrt d=0.11$ (Pythia-160m, initialisation), $0.24$ (160m, trained), $0.49$ (1b, trained) and $0.105$ (Qwen2.5-0.5B), against medians of order $0.1$--$4$ (Appendix~\ref{app:nondeg}). Qwen's blocks are sequential, so the input of the post-attention norm enters as well; its minimum is $0.012$ in the first block of the trained model and $0.14$ or more in all later blocks. For Qwen2.5-0.5B the constant of Assumption~\ref{ass:nondeg} is therefore $r/\sqrt d=0.012$, set by this first post-attention norm: the assumption holds, with a small margin in that block only. These checks are empirical: they cover the measured sequences and lengths and do not establish the assumption for arbitrary continuations or contexts. All other constants follow from the weights.

\subsection{Dilution}
\label{sec:theory-dilution}

Theorem~\ref{thm:nohair} rests on one geometric fact. A prompt of length $P$ occupies a fixed number of positions, while a query at distance $i$ sees $P+i$ keys; as $i$ grows, the prompt's share of the query's attention can only shrink. This is what we call dilution. We write $s_{ij}$ for the logit of query $i$ against key $j$, so that the attention weights are $a_{ij}=e^{s_{ij}}/\sum_ke^{s_{ik}}$. We first collect the constants that the architecture imposes (Lemma~\ref{lem:consts}), then make dilution precise (Lemma~\ref{lem:dilution}), and finally turn it into a recursion across layers (Lemma~\ref{lem:recursion}).

\begin{lemma}[Consequences of the architecture]\label{lem:consts}
Let $\|\cdot\|$ denote spectral norms of the weight matrices and $d_h$ the head dimension.
\begin{enumerate}
  \item[(a)] \emph{Bounded logits, unconditionally:} $|s_{ij}|\le B:=d\,\|W_Q\|\|W_K\|/\sqrt{d_h}$ (plus bias terms), since $\|\mathrm{LN}(h)\|=\sqrt d$ and rotary matrices are orthogonal.
  \item[(b)] \emph{LayerNorm is Lipschitz away from zero:} under Assumption~\ref{ass:nondeg}, $\|\mathrm{LN}(h)-\mathrm{LN}(h')\|\le (2\sqrt d/r)\,\|h-h'\|$.
  \item[(c)] Hence value-output vectors are bounded by $V:=\sqrt d\,\|W_OW_V\|$ and $\rho$-Lipschitz with $\rho:=2\sqrt d\,\|W_OW_V\|/r$; logits are $\lambda$-Lipschitz in each argument with $\lambda:=2d\,\|W_Q\|\|W_K\|/(r\sqrt{d_h})$; the MLP update is $\mu$-Lipschitz with $\mu:=\mathrm{Lip}(\sigma)\|W_2\|\|W_1\|\,2\sqrt d/r$, where $\sigma$ is the MLP nonlinearity, $W_1,W_2$ its weight matrices and $\mathrm{Lip}(\sigma)$ its Lipschitz constant.
\end{enumerate}
\end{lemma}
\begin{proof}
(a) The logit is $s_{ij}=\langle W_Q\mathrm{LN}(h_i),W_K\mathrm{LN}(h_j)\rangle/\sqrt{d_h}$. 
By Cauchy--Schwarz, we have: $|s_{ij}|\le\|W_Q\|\,\|\mathrm{LN}(h_i)\|\,\|W_K\|\,\|\mathrm{LN}(h_j)\|/\sqrt{d_h}$. Since $\|\mathrm{LN}(h)\|=\sqrt d$ and rotary matrices are orthogonal, this gives $|s_{ij}|\le d\,\|W_Q\|\|W_K\|/\sqrt{d_h}$.

(b) $\Pi$ is an orthogonal projection, hence non-expansive, so $\|\Pi u-\Pi v\|\le\|u-v\|$. For $\|u\|,\|v\|\ge r$,
\[
\Big\|\frac{\Pi u}{\|\Pi u\|}-\frac{\Pi v}{\|\Pi v\|}\Big\|
\le\frac{\|\Pi u-\Pi v\|}{\|\Pi u\|}+\|\Pi v\|\Big|\frac1{\|\Pi u\|}-\frac1{\|\Pi v\|}\Big|
\le\frac{\|\Pi u-\Pi v\|+|\|\Pi v\|-\|\Pi u\||}{r}
\le\frac{2\|u-v\|}{r},
\]
using $\|\Pi u\|,\|\Pi v\|\ge r$ in the second step and $\big|\|\Pi v\|-\|\Pi u\|\big|\le\|\Pi u-\Pi v\|\le\|u-v\|$ in the last.

(c) The value-output vector $W_OW_V\mathrm{LN}(h_j)$ has norm at most $\sqrt d\,\|W_OW_V\|=V$, and is $\rho$-Lipschitz with $\rho=2\sqrt d\,\|W_OW_V\|/r$, by (b). The logit is $\lambda$-Lipschitz in each argument with $\lambda=2d\,\|W_Q\|\|W_K\|/(r\sqrt{d_h})$, again by (b). The MLP update is $\mu$-Lipschitz with $\mu=\mathrm{Lip}(\sigma)\|W_2\|\|W_1\|\,2\sqrt d/r$. The constants are summed over heads in Appendix~\ref{app:proofs}.
\end{proof}

\begin{lemma}[Dilution]\label{lem:dilution}
Let $a_{ij}$ denote the attention weights of a head. For every query $i$ and every set $J$ of keys, $\sum_{j\in J}a_{ij}\le|J|\,e^{2B}/n_i$, where $n_i$ is the number of keys visible to the query. In particular, a continuation query at distance $i$ pays at most $Pe^{2B}/(P+i)\le Pe^{2B}/i$ to the prompt, in either run.
\end{lemma}
\begin{proof}
By definition, $a_{ij}=e^{s_{ij}}/\sum_k e^{s_{ik}}\le e^{s_{ij}-\min_ks_{ik}}/n_i\le e^{2B}/n_i$ and a prompt of length $P'\le P$ gives $P'/(P'+i)\le P/(P+i)$.
\end{proof}

\begin{lemma}[One-layer recursion]\label{lem:recursion}
Throughout, $V,\rho,\lambda V$ are the multi-head constants of Appendix~\ref{app:proofs}, already summed over heads. For $i\ge1$,
\begin{equation}\label{eq:recursion}
  \delta_l(i)\;\le\;K_1\,\delta_{l-1}(i)\;+\;\frac{K_2}{i}\sum_{j=1}^{i}\delta_{l-1}(j)\;+\;\frac{K_3\,P}{i},
\end{equation}
with $K_1=(1+\mu)(1+2\lambda V)$, $K_2=(1+\mu)\,e^{2B}(\rho+2\lambda V)$ and $K_3=3(1+\mu)\,e^{2B}V$ for a sequential block; for a parallel block the same bound holds with $K_1=(1+\mu)+2\lambda V$, which is no larger.
\end{lemma}
\begin{proof}
We give the main steps; the full proof, with the constants per head and the alignment of positions, is in Appendix~\ref{app:proofs}. Fix a head and a continuation query $i$, and let $u_j$, $u'_j$ be the value-output vectors of the two runs. The prompts may have different lengths, so we separate the prompt from the continuation, whose $i$ keys are the same in both runs and sit at the same relative positions. Let $\pi=\sum_{j\in\text{prompt}}a_{ij}$ and $\hat a_{ij}=a_{ij}/(1-\pi)$ for the continuation keys $j\le i$, and write the head output as $a=\pi\bar a_{\mathrm P}+(1-\pi)\bar a_{\mathrm C}$, with $\bar a_{\mathrm P}=\pi^{-1}\sum_{j\in\text{prompt}}a_{ij}u_j$ and $\bar a_{\mathrm C}=\sum_{j\le i}\hat a_{ij}u_j$; both have norm at most $V$, and similarly for the primed run. Then
\[
a-a'=(1-\pi)(\bar a_{\mathrm C}-\bar a'_{\mathrm C})+(\pi'-\pi)\,\bar a'_{\mathrm C}+\pi\,\bar a_{\mathrm P}-\pi'\,\bar a'_{\mathrm P}.
\]

\paragraph{Prompt terms.} By Lemma~\ref{lem:dilution}, $\pi,\pi'\le Pe^{2B}/i$, hence also $|\pi-\pi'|\le Pe^{2B}/i$, and the last three terms are at most $3V\,Pe^{2B}/i$.

\paragraph{Continuation term.} The weights $\hat a_{ij}$ are a softmax over the same $i$ keys in both runs, with logits in $[-B,B]$, so by Lemma~\ref{lem:softmax} in Appendix~\ref{app:proofs} we have $\hat a_{ij}\le e^{2B}/i$, also along the linear interpolation between the logits of the two runs. Write
\[
\bar a_{\mathrm C}-\bar a'_{\mathrm C}=\underbrace{\sum_j\hat a_{ij}(u_j-u'_j)}_{(\mathrm{i})}+\underbrace{\sum_j(\hat a_{ij}-\hat a'_{ij})u'_j}_{(\mathrm{ii})}.
\]
The first sum (i) is at most $\rho\sum_j\hat a_{ij}\delta_{l-1}(j)\le\rho e^{2B}i^{-1}\sum_{j\le i}\delta_{l-1}(j)$.

For the second sum (ii), the derivative of softmax gives $\|\hat a_i-\hat a'_i\|_1\le2\sup_{\text{path}}\sum_j\hat a_{ij}|s_{ij}-s'_{ij}|$ (Lemma~\ref{lem:softmax} in Appendix~\ref{app:proofs}), and $|s_{ij}-s'_{ij}|\le\lambda(\delta_{l-1}(i)+\delta_{l-1}(j))$, so (ii) is at most $2\lambda V\big(\delta_{l-1}(i)+e^{2B}i^{-1}\sum_{j\le i}\delta_{l-1}(j)\big)$.

Since $1-\pi\le1$, summing over heads (absorbed in the constants $V,\rho,\lambda V$) gives
\[
\|a-a'\|\le2\lambda V\delta_{l-1}(i)+e^{2B}(\rho+2\lambda V)\,i^{-1}\sum_{j\le i}\delta_{l-1}(j)+3Ve^{2B}P/i.
\]

\paragraph{Residual and MLP.} The residual adds $\delta_{l-1}(i)$. In a sequential block the MLP reads the post-attention residual, whose difference is at most $\delta_{l-1}(i)+\|a-a'\|$, and multiplies it by at most $1+\mu$, so $\delta_l(i)\le(1+\mu)\big(\delta_{l-1}(i)+\|a-a'\|\big)$. In a parallel block the MLP reads $h_{l-1}$ and the resulting bound $\delta_l(i)\le(1+\mu)\delta_{l-1}(i)+\|a-a'\|$ is smaller. Collecting terms gives \eqref{eq:recursion} with the constants stated, for the sequential case; the parallel case follows from the smaller $\delta_l(i)$ above.
\end{proof}

In words, \eqref{eq:recursion} says that the offset at layer $l$ and position $i$ is bounded by three contributions: the offset at the same position one layer earlier, the average offset over all earlier positions (the relay term), and the prompt's direct contribution, which is $O(P/i)$. Only the last term shrinks with $i$; the relay term is what carries prompt information through depth.
\subsection{The conditional no-hair theorem}
\label{sec:theory-main}
Now we give the following theorem where the no-hair analogy becomes precise.
\begin{theorem}[Conditional no-hair theorem]\label{thm:nohair}
Under Assumption~\ref{ass:nondeg}, for every $t\ge1$,
\begin{equation}\label{eq:nohair}
  \delta_L(t)\;\le\;C_L\,\frac{P\,\big(L+\ln t\big)^{L-1}}{t},
  \qquad C_L\;\le\;K_3\sum_{m=0}^{L-1}(K_1+2K_2)^{m}, 
\end{equation}
where we recall that $\delta_L(t)$ is the norm of the prompt offset at the final layer after $t$ generated tokens, that $P$ bounds the length of the two prompts, and that $L$ is the depth. The theorem bounds this offset by a constant $C_L$ that depends on the architecture but not on $t$ or $P$, and a decay in $t$ that is at worst $t^{-1}$ up to logarithmic factors.
\end{theorem}
\begin{proof}
We prove by induction on $l$ that $\delta_l(i)\le A_lP\,\ell(i)^{l-1}/i$, with $\ell(i)=L+\ln i\ge1$.

For $l=1$, $\delta_0\equiv0$, and \eqref{eq:recursion} gives $A_1=K_3$.

For $2\le l\le L$, put $k=l-2$. The function $f(s)=\ell(s)^k/s$ is decreasing on $[1,\infty)$, since $f'(s)=\ell(s)^{k-1}\big(k-\ell(s)\big)/s^2$ and $\ell(s)\ge L>k$. Hence
\[
\sum_{j\le i}f(j)\;\le\;f(1)+\int_1^if(s)\,ds\;\le\;L^{k}+\frac{\ell(i)^{k+1}}{k+1}\;\le\;\Big(\tfrac1L+\tfrac1{k+1}\Big)\ell(i)^{k+1}\;\le\;2\,\ell(i)^{l-1}.
\]
With $\ell(i)^{l-2}\le\ell(i)^{l-1}$ and $1\le\ell(i)^{l-1}$ for the other two terms, \eqref{eq:recursion} yields
\[
A_l\;\le\;(K_1+2K_2)\,A_{l-1}+K_3.
\]

Unrolling gives $C_L=A_L\le K_3\sum_{m=0}^{L-1}(K_1+2K_2)^m$. Since $K_1=(1+\mu)(1+2\lambda V)\ge1$, each term in the sum is at least $1$, so the sum is at most $L\,(K_1+2K_2)^{L-1}$ and $C_L\le L\,K_3\,(K_1+2K_2)^{L-1}$. Since $(L+\ln t)^{L-1}\sim(\ln t)^{L-1}$, the asymptotic form is $O\big(P(\ln t)^{L-1}/t\big)$.
\end{proof}

\begin{remark}[What the constant does and does not control]\label{rem:constant}
The constant is large and grows with depth. $C_L$ can grow exponentially with depth, and the bound is stated in terms of $e^{2B}$, which is large. Lemma~\ref{lem:dilution} only uses the spread $\Delta=\max_ks_{ik}-\min_ks_{ik}$ of the logits seen by each query, so $e^{2B}$ can be replaced by $e^{\Delta}$; but in trained networks with attention sinks $\Delta$ is large, and the gain is limited.
Intuitively, relay acts as a memory buffer: the prompt's information is carried forward by the generated text itself, but it is subject to repeated attenuation and mixing at each layer.

What Theorem~\ref{thm:nohair} fixes is the asymptotic shape, not a numerical prediction. The bound is informative only once \(t\gg Pe^{\Delta}\), which for trained networks lies beyond any window we measure; the fitted exponents of trained models, which are below $1$, are therefore neither in conflict with the theorem nor predicted by it. What the theorem and the solvable model establish together is an asymptotic statement: no decay is slower than \(t^{-1}\) up to at most \(L-1\) powers of \(\ln t\), one per relay round, and for uniform attention this cannot be improved. The shift \(L\) inside the logarithm only matters for small \(t\). For the finite windows of the untrained networks, the relevant description is the exactly solvable linear relay (Proposition~\ref{prop:gains}); for trained networks we have no corresponding quantitative model, and their exponents are measurements (Section~\ref{sec:results-alpha}).

How large $\Delta$ can be is observed in practice. For Pythia-160m we measure, over late queries of 400 prompt--continuation pairs, the spread of each query's logits, $\ln(\max_k\alpha_{ik}/\min_k\alpha_{ik})$ with $\alpha_{ik}=e^{s_{ik}}$. At initialisation its median is $1.9$--$2.3$ (maximum $4.3$), so $e^{\Delta}$ is of order $10^2$ and the dilution regime is reached within our windows. After training the median is $17$--$18$ in layers 1--3, so $e^{\Delta}\gtrsim10^{7}$; in the deep layers the spread reaches the float32 limit ($\Delta\ge69$, where the smallest attention weights underflow), so there it is only bounded from below. In all these layers the bound is not informative at any accessible $t$.
\end{remark}
\begin{remark}[From $\delta$ to the measured signal]\label{rem:jensen}
Since $\|\mathbb E_c\Delta\|\le\mathbb E_c\|\Delta\|$, the signal of Section~\ref{sec:measure} obeys $\mathcal S_l(t)\le\big(\mathbb E_c\,\delta_l(t)\big)^2/\big(d\,\min_j\sigma_{l,j}^2\big)$, so Theorem~\ref{thm:nohair} bounds it as well (averaging over a window only moves $t$ within the window). The converse fails: offsets whose direction depends on the continuation partly cancel in $\mathbb E_c\Delta$, so $\mathcal S$ can decay faster than $\delta^2$. We therefore also report the exponent of $\mathcal M_l$ (Section~\ref{sec:measure-est}), which is the window-averaged $\mathbb E_c\delta_l^2$ up to the coordinate scaling.
\end{remark}

\subsection{The exterior carries the hair}
\label{sec:theory-exterior}

\begin{corollary}[The exterior carries the hair]\label{cor:exterior}
Let $p(\cdot\mid x,c_{<t})$ be the next-token distribution given the history $c_{<t}$ (the prefix of the continuation up to time $t$), $W_U$ the unembedding, and $\varepsilon_t=\|W_U\|_{2\to\infty}\,(2\sqrt d/r)\,C_L\,P\,(L+\ln t)^{L-1}/t$, where $\|W_U\|_{2\to\infty}$ is the operator norm from $\ell_2$ to $\ell_\infty$ and $\mathrm{TV}$ the total-variation distance. For every history $c_{<t}$,
\[
\mathrm{TV}\big(p(\cdot\mid x,c_{<t}),\,p(\cdot\mid x',c_{<t})\big)\le\varepsilon_t=O\big(P(\ln t)^{L-1}/t\big),
\]
and for every $m\ge0$ the joint laws of the next $m+1$ tokens given $c_{<t}$ differ in total variation by at most $\sum_{s=t}^{t+m}\varepsilon_s\le(m+1)\,\varepsilon_t$.
\end{corollary}
\begin{proof}
The final LayerNorm is Lipschitz by Lemma~\ref{lem:consts}(b), and softmax is $1$-Lipschitz from $\ell_\infty$ logits to total variation; Theorem~\ref{thm:nohair} holds for every continuation, hence for every history. For the joint law, couple the two runs token by token, maximally at each step given a common history: the probability that the coupled sequences first differ at step $s$ is at most $\varepsilon_s$. Finally $\varepsilon_s$ is non-increasing in $s$, by the monotonicity argument in the proof of Theorem~\ref{thm:nohair}.
\end{proof}

The corollary says what happens over a short window. Over a window short compared with $t/\big(P(\ln t)^{L-1}\big)$, the next token is determined, up to $(m+1)\varepsilon_t$, by the text generated so far, whichever prompt produced it.

It says nothing about long horizons. The bound applies step by step; the sum $\sum_s\varepsilon_s$ diverges, so it does not exclude a cumulative influence of the prompt that bypasses the text. Nor does it show that such an influence exists.

Whether the prompt's influence at late times is in fact carried by the text is therefore an empirical statement, which Section~\ref{sec:exterior} tests, not a consequence of Theorem~\ref{thm:nohair}.

This is the sense in which the no-hair theorem holds. In general relativity the quantities that survive are those measurable at infinity, carried by the exterior field; here, step by step, the channel through which a prompt influences what comes next is the exterior record. Section~\ref{sec:exterior} is consistent with this picture: the language of the prompt remains decodable from free generations, but under a shared continuation its imprint in the hidden state still decays, at a rate comparable to that of a one-word change.

\subsection{An exactly solvable model}
\label{sec:theory-linear}

Theorem~\ref{thm:nohair} gives an upper bound, and the bound contains logarithmic powers. Are they an artefact of the proof, or a feature of the architecture? To answer this, we turn to a simplified model that can be solved exactly,  which is, the \emph{linear relay model}. In this model,  every head attends uniformly, each block multiplies the offset by a constant gain, and the prompt enters uniformly. The model is solvable in closed form, and its solution has the same $P(\ln i)^{l-1}/i$ shape as the bound, which shows that the logarithmic powers cannot be removed. It also gives a parameter-free prediction for untrained networks, which we compare with the measured curves at the end of this subsection.

\paragraph{The model.} Consider the first-order dynamics of a small offset around a fixed background: every head attends uniformly ($a_{ij}=1/n_i$, the regime of a randomly initialised network, whose logits are close to zero), the value-output path of block $l$ multiplies the offset by a gain $g_l>0$ along a fixed direction $e$, and normalisation and MLPs are replaced by their Jacobians, which we absorb into $g_l$. The two prompts differ by $e$ at every prompt position, and we keep this prompt offset fixed through depth (if it grows, every offset below grows, so lower bounds remain valid). The offset at the $i$-th continuation token after $l$ blocks is then $\delta_l(i)\,e$ with
\begin{equation}\label{eq:linrelay}
  \delta_l(i)=\delta_{l-1}(i)+\frac{g_l}{P+i}\Big(P+\sum_{j=1}^{i}\delta_{l-1}(j)\Big),\qquad\delta_0\equiv0 .
\end{equation}

 The first result says below that the bound from Theorem~\ref{thm:nohair} is attained in this regime: with equal gains, the offset decays with exactly $l-1$ logarithmic powers, so the exponent cannot be lowered.

\begin{proposition}[Uniform attention attains the bound]\label{prop:tight}
In the linear relay model with equal gains $g$, for all $l,i\ge1$,
\begin{equation}\label{eq:tight}
  \frac{g^{l}}{(l-1)!}\,\frac{P\,\big(\ln\frac{P+i+1}{P+1}\big)^{l-1}}{P+i}\;\le\;\delta_l(i)\;\le\;C_l\,\frac{P\,(l+\ln i)^{l-1}}{i},
  \qquad C_l\le g\sum_{m=0}^{l-1}(1+2g)^m .
\end{equation}
Hence $\delta_l(i)=\Theta\big(P(\ln i)^{l-1}/i\big)$ as $i\to\infty$: the power $l-1$ in Theorem~\ref{thm:nohair} cannot be lowered.
\end{proposition}
\begin{proof}
Since $P+i\ge i$, \eqref{eq:linrelay} is of the form \eqref{eq:recursion} with $K_1=1$ and $K_2=K_3=g$, and the induction in the proof of Theorem~\ref{thm:nohair} applies with $L$ replaced by $l$. The lower bound is the $k=l$ term of the exact solution \eqref{eq:gains} below, with $e_l=g^l$ and $h_{l-1}\ge p_1^{l-1}/(l-1)!\ge\big(\ln\frac{P+i+1}{P+1}\big)^{l-1}/(l-1)!$; all other terms are non-negative.
\end{proof}

\paragraph{Layer-dependent gains.} The model can also be solved exactly for arbitrary gains, which turns the depth dependence of untrained networks into a prediction.

\begin{proposition}[Exact solution with layer gains]\label{prop:gains}
Let $x_j=1/(P+j)$. We write $e_k$ and $h_k$ for the elementary and the complete homogeneous symmetric polynomials in $g_1,\dots,g_l$ and in $x_1,\dots,x_i$ ($e_0=h_0=1$, $h_{-1}=0$). In the linear relay model, for all $l,i\ge1$,
\begin{equation}\label{eq:gains}
  \delta_l(i)\;=\;\frac{P}{P+i}\sum_{k=1}^{l}e_k(g_1,\dots,g_l)\,h_{k-1}(x_1,\dots,x_i).
\end{equation}
Moreover, with $u=\ln\frac{P+i}{P}$, for every $k\ge1$
\begin{equation}\label{eq:hbound}
  \Big|h_k(x_1,\dots,x_i)-\frac{u^k}{k!}\Big|\;\le\;\frac1P\Big(\frac{u^{k-1}}{(k-1)!}+h_{k-2}(x_1,\dots,x_i)\Big),
\end{equation}
so the continuum form $\delta_l(i)\simeq\frac{P}{P+i}\sum_ke_k\,\frac{u^{k-1}}{(k-1)!}$ holds up to corrections of order $1/P$ times terms of lower degree in $u$. For equal gains $e_k=\binom lkg^k$ and the leading power is $u^{l-1}$.
\end{proposition}
\begin{proof}
We first derive \eqref{eq:gains}, then the bound \eqref{eq:hbound}.

\emph{Deriving \eqref{eq:gains}.} Put
\[
\varphi_l(i)=\frac{P+i}{P}\delta_l(i),\qquad
(Jf)(i)=\sum_{j\le i}x_jf(j),\qquad
\chi_l=1+J\varphi_l,
\]
so that $\chi_l(i)-\chi_l(i-1)=x_i\varphi_l(i)$ and $\chi_l(0)=1$. Since $\delta_{l-1}(j)/P=x_j\varphi_{l-1}(j)$, \eqref{eq:linrelay} reads
\[
\varphi_l=\varphi_{l-1}+g_l\chi_{l-1},\qquad\varphi_0=0,\ \chi_0=1.
\]
Applying $J$ and adding $1$ gives $\chi_l=(1+g_lJ)\chi_{l-1}$, hence
\[
\chi_l=\prod_{m\le l}(1+g_mJ)\,1=\sum_{k}e_k(g_1,\dots,g_l)\,J^k1.
\]
By induction on $k$, $(J^k1)(i)=\sum_{j_1\le\dots\le j_k\le i}x_{j_1}\cdots x_{j_k}=h_k(x_1,\dots,x_i)$, and $h_k(x_1,\dots,x_i)-h_k(x_1,\dots,x_{i-1})=x_i\,h_{k-1}(x_1,\dots,x_i)$. Dividing $\chi_l(i)-\chi_l(i-1)$ by $x_i$ gives \eqref{eq:gains}.

\emph{Deriving \eqref{eq:hbound}.} Let $p_1=\sum_{j\le i}x_j$ and $p_2=\sum_{j\le i}x_j^2<1/P$. Over multisets $m$ of size $k$, $h_k=\sum_mx^m$ and $p_1^k/k!=\sum_mx^m/m!$ with $m!=\prod_rm_r!$. The difference is non-negative and involves only multisets with a repeated index $r$, whose monomials appear in $x_r^2h_{k-2}$, hence
\[
0\le h_k-\frac{p_1^k}{k!}\le p_2\,h_{k-2}.
\]
Comparing $p_1$ with $\int ds/s$ gives $\ln\frac{P+i+1}{P+1}\le p_1\le u$, so
\[
0\le u-p_1\le\ln\Big(1+\frac{i}{P(P+i+1)}\Big)<\frac1P
\quad\text{and}\quad
0\le\frac{u^k-p_1^k}{k!}\le\frac{u^{k-1}}{P\,(k-1)!}.
\]
Combining the two estimates gives \eqref{eq:hbound}.
\end{proof}

\paragraph{Small gains.} When the gains are small, the exact solution reduces to a single-parameter form and gives an explicit expression for the local exponent.

\begin{corollary}[Small gains]\label{cor:smallgain}
Let $G_l=\sum_{m\le l}g_m$ and $Q_l=\sum_{m\le l}g_m^2$. Then $0\le G_l^k/k!-e_k\le\frac{Q_l}{2}\,G_l^{k-2}/(k-2)!$. If $Q_lu\ll G_l$, $G_l\ll P$ and $l^2\gg G_lu$ (the three error terms of Corollary~\ref{cor:smallgain-quant}; note that $Q_l\ll G_l^2$ alone does not suffice once $G_lu$ is large), then
\begin{equation}\label{eq:bessel0}
  \delta_l(i)\simeq\frac{P}{P+i}\,\varphi(G_l,u),\qquad
  \varphi(G,u)=\sum_{k\ge1}\frac{G^ku^{k-1}}{k!\,(k-1)!}=\sqrt{G/u}\;I_1\big(2\sqrt{Gu}\big),
\end{equation}
with $I_1$ a modified Bessel function, and the local exponent $\alpha_{\mathrm{loc}}=-d\ln\delta_l/d\ln t$ is
\begin{equation}\label{eq:bessel}
  \alpha_{\mathrm{loc}}\simeq\frac{t}{P+t}\big(1-\partial_u\ln\varphi\big),\qquad
  \partial_u\ln\varphi=\sqrt{G_l/u}-\frac{3}{4u}+O\big((G_lu^3)^{-1/2}\big)\quad(G_lu\gg1).
\end{equation}
Depth enters only through the accumulated gain $G_l$; Proposition~\ref{prop:sandwich} and Corollary~\ref{cor:smallgain-quant} in Appendix~\ref{app:proofs} bound the error of \eqref{eq:bessel0} explicitly. Moreover (a) $\alpha_{\mathrm{loc}}\to1$ as $t\to\infty$, and (b) at fixed $t$, $\alpha_{\mathrm{loc}}$ does not increase as $G_l$ grows, hence decreases with depth. Over a finite window the measured exponent of an untrained network is therefore below $1$, lower in deeper layers, and creeps up as the window is extended.
\end{corollary}
\begin{proof}
For the bound on $e_k$, write $G^k/k!=\sum_mg^m/m!$ over multisets of size $k$ and $e_k$ as the sum over those without repeated index; a multiset with a repeated index $r$ contributes at most $\tfrac{g_r^2}{2}$ times a term of $G^{k-2}/(k-2)!$. Inserting $e_k\simeq G^k/k!$ and $h_{k-1}\simeq u^{k-1}/(k-1)!$ (Proposition~\ref{prop:gains}) into \eqref{eq:gains} gives \eqref{eq:bessel0}; the terms of the series peak at $k\approx\sqrt{Gu}$, so the truncation at $k=l$ is immaterial. Since $du/d\ln t=t/(P+t)$, $\alpha_{\mathrm{loc}}=\frac{t}{P+t}(1-\partial_u\ln\varphi)$, and the expansion follows from $I_1(z)=e^z(2\pi z)^{-1/2}(1+O(1/z))$. For (a), $\partial_u\ln\varphi\to0$. For (b), write $\varphi=G\,F(Gu)$ with $F(z)=\sum_kz^{k-1}/(k!(k-1)!)$, so $\partial_u\ln\varphi=u^{-1}\zeta(Gu)$ with $\zeta(z)=zF'(z)/F(z)$. Now $\zeta(z)$ is the mean of $k-1$ under weights proportional to $z^{k-1}/(k!(k-1)!)$, and its derivative in $\ln z$ is the corresponding variance, so $\zeta$ is non-decreasing and $\partial_u\ln\varphi$ is non-decreasing in $G$.
\end{proof}

\paragraph{Comparison with experiment.} The model makes two parameter-free predictions for the first blocks. In layer 1, $\delta_1=g_1P/(P+i)$, so the local exponent is exactly $t/(P+t)$. In layer 2, $\varphi_2=(g_1+g_2)+g_1g_2h_1$, so $\big(1-\tfrac{P+t}{t}\alpha_{\mathrm{loc}}\big)^{-1}=h_1+\tfrac{g_1+g_2}{g_1g_2}$ is linear in $h_1\approx u$ with slope exactly $1$, whatever the gains. Untrained Pythia is consistent with both (Section~\ref{sec:results-alpha}: the layer-1 law with $P_{\mathrm{eff}}$ close to the prompt length, fitted power $1.07$--$1.33$ in layer 2); untrained Qwen is consistent with neither, so its first blocks are not in the linear-relay regime. In deeper layers the effective power seen over a finite window depends on the gains, not only on the layer index, so the coincidence $k\approx l-1$ in layers 2--4 of Pythia should not be read as a general law.

\emph{Small-gain fit.} Fitting \eqref{eq:bessel} without the factor $t/(P+t)$ and the $3/(4u)$ term to untrained Pythia-1b on real-corpus pairs ($P\approx28$, unbiased signal, centre-labelled windows) gives $R^2\ge0.99$ in layers 3--15, with $G_l=0.36,0.52,0.95,1.20,1.35$ in layers $3,4,8,12,15$ and increments of roughly $0.65/l$ per layer, i.e.\ $G_l\approx0.65\ln l+\mathrm{const}$. Repeating the fit in all fifteen layers gives $G_l=0.60\ln l-0.29$ ($R^2=0.997$, largest residual $0.05$) and per-layer increments $g_l\propto l^{-0.94}$; the fitted logarithmic power follows $k=2.09\ln l-0.12$ ($R^2=0.9998$). Over our windows, however, $G_lu\le5$ (and $\le1.3$ in layer 3), outside the range of \eqref{eq:bessel}, so these values are effective parameters rather than estimates of the accumulated gain.

\emph{Exact-solution fit.} The exact solution \eqref{eq:gains} turns them into estimates. We fit it to the same curves layer by layer, with the exact recursion and one free amplitude per layer (the coordinate normalisation of $\mathcal S$ is layer dependent), so that each layer is fitted on the shape of its curve only. Layer 1 fixes $P$ whatever the gains; it gives $P=28.8$, against a mean prompt length of $25.7$ tokens, with $R^2=0.9999$ and a largest deviation of the local exponent of $0.026$. Since $\delta_l$ depends on the gains only through the symmetric functions $e_k(g_1,\dots,g_l)$, layer 2 constrains only $g_1g_2/(g_1+g_2)$; we fit $g_1,g_2,g_3$ jointly on layers 2--3 and then, from layer 4 on, one new gain per layer with the earlier ones fixed, so that a single parameter predicts the whole curve of each layer. All fifteen curves are reproduced with $R^2\ge0.992$; the largest deviation between measured and predicted local exponents grows from $0.05$ in layer 2 to $0.08$ in layer 15, so the one-gain-per-layer model captures the curves well but not perfectly in the deep layers.

\emph{Identifying the first two gains.} Layer 2 alone constrains $g_1$ and $g_2$ only through $g_1g_2/(g_1+g_2)$; together with layer 3 they are determined as an unordered pair, since every $\delta_l$ is symmetric in $g_1,\dots,g_l$, so $g_1$ and $g_2$ cannot be told apart. The fit returns equal values, $\hat g_1=\hat g_2=0.66$.

\emph{A test of the symmetry.} This is not an artefact of the starting point: restarting the joint fit of $g_1,g_2,g_3$ from six widely different initial values, including strongly asymmetric ones (e.g.\ $(0.05,2,0.2)$ and $(2,0.2,0.5)$), reaches the same minimum $(\hat g_1,\hat g_2,\hat g_3)=(0.664,0.664,0.24)$ at the same cost each time, and on synthetic curves generated with an unequal pair $(0.65,0.325)$ the same fit, started on the symmetric diagonal, leaves it and recovers the unequal values, so the equality is a property of the data and not of the optimiser.

\emph{The remaining gains.} The gains then decrease as $\hat g_l\propto l^{-1.4}$ (layers 4--15), to $0.02$ in layer 15, with $\hat G_{15}=2.18$; a joint fit of all gains gives the same values to within $0.09$. The first gains are not small, which is why the effective $G_l$ of the small-gain fit differ from these estimates.

\begin{remark}[Running exponent]\label{rem:running}
For $t\gg P$ and $G_lu\gg1$, Eq.~\eqref{eq:bessel} gives $\alpha_{\mathrm{loc}}\simeq1-\sqrt{G_l/u}+3/(4u)$ with $u\simeq\ln(t/P)$. The approach to the asymptotic exponent $1$ is therefore logarithmically slow, with leading correction $(\ln t)^{-1/2}$ whose amplitude $\sqrt{G_l}$ grows with depth. In the language of critical phenomena, $\alpha_{\mathrm{loc}}(t)$ is a running exponent, and the accumulated gain $G_l$, which our fits give as growing roughly like $\ln l$ with the layer index (an effective parameter, see above), acts as a depth-dependent coupling that sets the crossover scale $u^\ast\sim1/G_l$ below which the decay is not yet in its asymptotic regime. Over our windows, $G_lu\le5$, so the untrained networks stay in this crossover regime.
\end{remark}

\begin{remark}[Physics analogy]
Lemma~\ref{lem:dilution} is a Gauss-law statement: a fixed charge (here the prompt, a fixed number $P$ of positions) spread over a growing volume (the $P+i$ visible keys), so that its share of any query's attention falls as $1/i$. Relay plays the role of back-scattering off the background curvature, which in general relativity produces late-time power-law tails \citep{price1972}: a perturbation that has left the region of direct view returns through intermediate positions, and in the linear relay model each relay round contributes one factor of $\ln t$. The analogy is structural, not quantitative. We make no claim that the exponent or the amplitude of the tail corresponds to a quantity of the black-hole problem, and the identification of a prompt with a conserved charge is only suggestive, since a transformer has no conservation law that protects any feature of its prompt. Trained networks scatter little; untrained ones scatter a lot. We use the analogy as a guide only, and no result of this paper depends on it.
\end{remark}
\section{Forgetting is close to a power law}
\label{sec:results-alpha}

We now turn to the measurement. The theory of Section~\ref{sec:theory} predicts the shape of the decay but not its rate; the rate is the exponent $\alpha_l$ defined in Section~\ref{sec:measure}, and this section reports the corresponding results.  
\paperfig{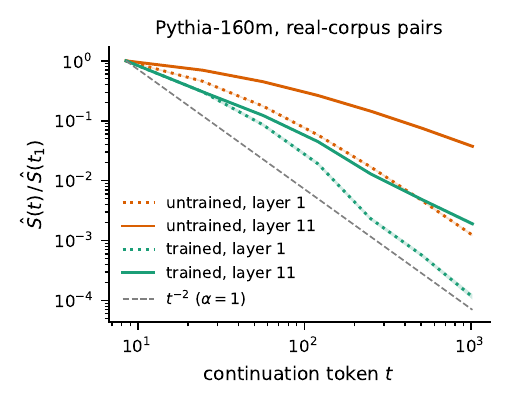}{0.5}{Decay of the prompt offset under a shared continuation (Pythia-160m, real-corpus pairs, unbiased signal normalised to the first window; bands: 95\% bootstrap over pairs). Dotted: layer 1; solid: last layer. Training makes the deep-layer offset decay much faster; the dashed line is $\alpha=1$.}{fig:decay}

\subsection{Hand-written scenes}
\label{sec:results-hand}
Table~\ref{tab:alpha} gives the exponent $\alpha$ for Pythia-160m and 410m at initialisation and after training. Five patterns are visible, and they carry over to the real-corpus pairs of Section~\ref{sec:results-real}.
\begin{table}[t]
\centering\small
\begin{tabular}{lcccccc}
\toprule
\multicolumn{7}{l}{\emph{Pythia-160m (12 layers)}}\\
 & L1 & L2 & L3 & L6 & L9 & L11 \\
\midrule
init (step 0)   & 0.82 {\scriptsize[.81,.82]} & 0.65 {\scriptsize[.65,.65]} & 0.57 {\scriptsize[.57,.58]} & 0.47 {\scriptsize[.47,.47]} & 0.42 {\scriptsize[.42,.43]} & 0.40 {\scriptsize[.40,.41]} \\
trained (143k)  & 1.05 {\scriptsize[1.04,1.07]} & 0.92 {\scriptsize[.90,.93]} & 0.63 {\scriptsize[.57,.71]} & 0.59 {\scriptsize[.55,.62]} & 0.64 {\scriptsize[.62,.67]} & 0.66 {\scriptsize[.64,.68]} \\
\midrule
\multicolumn{7}{l}{\emph{Pythia-410m (24 layers)}}\\
 & L1 & L2 & L6 & L12 & L18 & L23 \\
\midrule
init (step 0)   & 0.82 {\scriptsize[.81,.82]} & 0.65 {\scriptsize[.65,.66]} & 0.47 {\scriptsize[.47,.48]} & 0.40 {\scriptsize[.39,.40]} & 0.36 {\scriptsize[.36,.37]} & 0.35 {\scriptsize[.34,.35]} \\
trained (143k)  & 1.01 {\scriptsize[1.00,1.03]} & 0.89 {\scriptsize[.88,.90]} & 0.70 {\scriptsize[.69,.72]} & 0.57 {\scriptsize[.56,.58]} & 0.60 {\scriptsize[.59,.61]} & 0.60 {\scriptsize[.59,.61]} \\
\bottomrule
\end{tabular}
\caption{Decay exponent $\alpha$ of the prompt offset, hand-written scenes (20 scenes, 16 continuations each), $t_e\in[16,1920]$, windows labelled by their centre, plug-in signal (not the unbiased estimator of Section~\ref{sec:measure-est}: late-time exponents are biased low, so the unbiased values of Tables~\ref{tab:real} and~\ref{tab:scale} are the reference), 95\% bootstrap intervals over scenes. $|\beta|\le0.07$ in all cells.}
\label{tab:alpha}
\end{table}

\begin{itemize}
  \item \textbf{Noise does not decay.} $\beta\approx 0$ for every Pythia model, layer and checkpoint (untrained Qwen is the exception, Section~\ref{sec:results-qwen}), so all the change in decodability comes from $\alpha$. Since $\mathcal N$ is the spread of hidden states across continuations, this is what stationarity of the continuation text implies; we report it as a check rather than as a finding.
  \item \textbf{Layer 1 is pure dilution.} After the first attention block the continuation tokens are still identical in the two runs, so no relay is possible and only the direct path exists. In untrained models the local exponent follows the dilution law of a uniform head, $\ln\frac{P+t_2}{P+t_1}\big/\ln\frac{t_2}{t_1}$ between consecutive points, with a single fitted $P_{\mathrm{eff}}$ to within $0.03$ at every octave; the fitted value tracks the prompt length ($P_{\mathrm{eff}}\approx17$ for the hand-written scenes, $\approx30$ for the longer real-corpus prompts). Trained models give $\alpha\approx1$ in layer 1 ($1.05$ and $1.01$ here, $0.94$--$0.97$ on real-corpus pairs).
  \item \textbf{Relay enters after layer 2 in trained models.} Layers 1--2 stay at $0.9$--$1.05$; deeper layers settle at $0.57$--$0.70$ (both sizes) and are roughly flat with depth.
  \item \textbf{At initialisation, $\alpha$ depends on the layer index, not on width.} At matched layers the two untrained models agree exactly (L1 $0.82$/$0.82$, L2 $0.65$/$0.65$, L6 $0.47$/$0.47$). In layer 1 this is automatic: pure dilution depends only on the prompt length, which is the same for both models (same tokenizer). The agreement in deeper layers is informative: in the linear relay model (Proposition~\ref{prop:gains}) it means that the gains at a given layer index do not depend on the width. This is not because the initial weights match. The released step-0 checkpoints follow small\_init for the input projections (standard deviation $\sqrt{2/5d}$: $0.0228$, $0.0198$, $0.0140$ for 160m, 410m, 1b) and wang\_init for the output projections ($2/(L\sqrt d)$: $0.0060$, $0.0026$, $0.0028$), so the absolute size of each block's write differs between the three models and depends on their total depth. A scale-free mechanism is needed. A natural candidate is that what matters is the write of a block relative to the residual stream it adds to, which at initialisation is itself made of the earlier writes: the pre-norm activations grow as $\sqrt l$ with depth (Pythia-160m: $\|\Pi h\|/\sqrt{d}\approx0.12\sqrt l$ to within $0.01$ from block 1 to block 11), so the relative write of block $l$, and with it the gain, depends on $l$ but not on the absolute scale.  
  \item \textbf{Content-agnostic vs.\ content-dependent.} At initialisation the axes have the same $\alpha$ to within $0.04$; after training they separate, in the same order for both sizes (deep layers on hand-written scenes: paraphrase $0.50$--$0.57$, detail $0.65$--$0.75$).
\end{itemize}

\paragraph{Local exponents rise towards $1$.} If $\delta_l(t)\propto(\ln t)^{k}/t$, the local exponent is $\alpha_{\mathrm{loc}}=1-k/\ln(t/t_0)$ for some reference time $t_0$. Then $1/(1-\alpha_{\mathrm{loc}})$ is linear in $\ln t$ with slope $1/k$, and we use $k$ as a descriptive effective power. For untrained networks $\alpha_{\mathrm{loc}}$ rises monotonically with $t$ in every layer, as Corollary~\ref{cor:smallgain} predicts: in Pythia-160m, from $0.47$ to $0.99$ in layer 1, and from $0.21$ to $0.53$ in layer 11. The linear form fits with $R^2\ge0.985$ (Table~\ref{tab:keff}). In layer 2 the fitted power is close to $1$ ($1.07$--$1.08$), the value the linear relay model gives for any gains (Section~\ref{sec:theory-linear}, \emph{Comparison with experiment}). In layer 3 it is close to $2$ ($1.96$), and from layer 6 on it grows much more slowly than $l-1$ ($k/\ln l\approx2.0$--$2.2$ in both models), as expected when the gains are small. These fits ignore the factor $t/(P+t)$ of \eqref{eq:bessel}; the exact solution \eqref{eq:gains} is the appropriate model for the full curves. The small-gain form of Corollary~\ref{cor:smallgain} fits the same local exponents at least as well, and we compare the two on the real-corpus pairs, where the unbiased signal is available.

\begin{table}[t]
\centering\small
\begin{tabular}{lccccc}
\toprule
\multicolumn{6}{l}{\emph{Pythia-160m, hand-written scenes}}\\
layer $l$ & 2 & 3 & 6 & 9 & 11 \\
\midrule
fitted $k$ & 1.07 & 1.96 & 3.59 & 4.65 & 5.10 \\
relay rounds $l-1$ & 1 & 2 & 5 & 8 & 10 \\
$R^2$ & 0.985 & 0.996 & 0.995 & 0.993 & 0.996 \\
\midrule
\multicolumn{6}{l}{\emph{Pythia-410m, hand-written scenes}}\\
layer $l$ & 2 & 6 & 12 & 18 & 23 \\
\midrule
fitted $k$ & 1.08 & 3.54 & 5.26 & 6.21 & 6.76 \\
relay rounds $l-1$ & 1 & 5 & 11 & 17 & 22 \\
$R^2$ & 0.988 & 0.996 & 0.995 & 0.997 & 0.996 \\
\midrule
\multicolumn{6}{l}{\emph{Pythia-1b, real-corpus pairs (unbiased signal)}}\\
layer $l$ & 2 & 3 & 4 & 8 & 15 \\
\midrule
fitted $k$ & 1.33 & 2.14 & 2.78 & 4.24 & 5.52 \\
$R^2$ & 0.969 & 0.986 & 0.994 & 0.996 & 0.997 \\
fitted $G_l$ (Cor.~\ref{cor:smallgain}) & 0.17 & 0.36 & 0.52 & 0.95 & 1.35 \\
$R^2$ & 0.966 & 0.990 & 0.999 & 0.998 & 0.996 \\
\bottomrule
\end{tabular}
\caption{Effective logarithmic power $k$ and accumulated gain $G_l$, fitted from the local exponents of the untrained models (windows labelled by their centre). The $G_l$ are effective parameters of the small-gain form; the estimates from the exact solution, which differ because the first gains are not small, are in Section~\ref{sec:theory-linear}.}
\label{tab:keff}
\end{table}

\paperfig{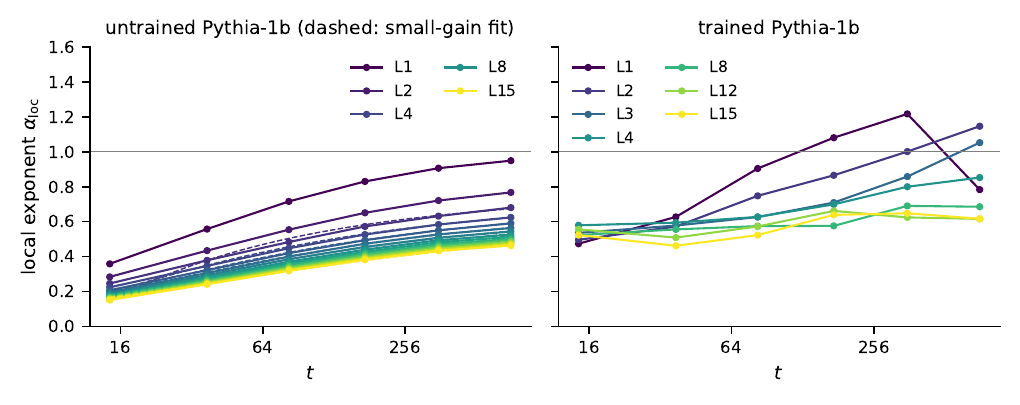}{1.0}{Local exponents between consecutive octaves. Left: every layer of untrained Pythia-1b; the exponent rises towards $1$ in every layer and decreases with depth; dashed: the small-gain form of Corollary~\ref{cor:smallgain}, fitted as $(1-\alpha_{\mathrm{loc}})^{-2}=u/G_l+c$ (the offset $c$ absorbs the finite-$u$ corrections, Section~\ref{sec:theory-linear}). Right: the trained model, where no systematic trend remains in the deep layers.}{fig:local}

\subsection{Real-corpus prompt pairs}
\label{sec:results-real}

To rule out an artefact of hand-written scenes we repeat the measurement on 100 pairs per axis drawn from real corpora: a WikiText-103 test sentence vs.\ the same sentence with one content word swapped (\emph{detail}); MRPC paraphrase pairs (\emph{para}); WMT14 English sentences vs.\ their French translations (\emph{lang}); and two random WikiText sentences (\emph{unrelated}), the largest possible difference. Continuations are WikiText-103 validation chunks, $t_e\in[16,1024]$, 8 continuations per pair, unbiased signal; intervals are over prompt pairs (Table~\ref{tab:real}).

\begin{table}[t]
\centering\small
\begin{tabular}{lcccccccc}
\toprule
 & \multicolumn{4}{c}{init (step 0)} & \multicolumn{4}{c}{trained (step 143000)} \\
\cmidrule(lr){2-5}\cmidrule(lr){6-9}
layer & detail & para & lang & unrel. & detail & para & lang & unrel. \\
\midrule
L1  & 0.75 & 0.74 & 0.69 & 0.73 & 0.92 & 0.98 & 0.99 & 0.97 \\
L2  & 0.59 & 0.58 & 0.54 & 0.57 & 0.80 & 0.81 & 0.80 & 0.82 \\
L3  & 0.52 & 0.51 & 0.47 & 0.50 & 0.67 & 0.52 & 0.56 & 0.67 \\
L6  & 0.42 & 0.42 & 0.39 & 0.41 & 0.61 & 0.54 & 0.54 & 0.63 \\
L9  & 0.39 & 0.38 & 0.35 & 0.37 & 0.67 & 0.58 & 0.61 & 0.67 \\
L11 & 0.37 & 0.37 & 0.33 & 0.35 & 0.69 & 0.59 & 0.65 & 0.70 \\
\bottomrule
\end{tabular}
\caption{$\alpha$ on real-corpus prompt pairs, Pythia-160m, 100 pairs per axis, unbiased signal, windows labelled by their centre. 95\% intervals over pairs have half-width $\le0.02$ at initialisation and $\le0.06$ after training; $|\beta|\le0.08$.}
\label{tab:real}
\end{table}

Every finding carries over. At initialisation $\alpha$ falls with depth and the local exponent rises monotonically with $t$ in every layer; the fitted powers are $k=1.33, 2.11$ in layers 2 and 3 ($R^2\ge0.97$), then $k/\ln l\approx2.0$; the layer-1 local exponent follows the dilution law with $P_{\mathrm{eff}}\approx30$, close to the length of these prompts (26--32 tokens on average), against $\approx17$ for the shorter hand-written ones: the prompt's share of the context scales with its length. After training, layers 1--2 stay at $0.8$--$1$ and deeper layers at $0.52$--$0.70$, with paraphrase and language forgotten most slowly.

\paragraph{Two descriptions of the untrained curves.} Corollary~\ref{cor:smallgain} predicts that $(1-\alpha_{\mathrm{loc}})^{-2}$ is linear in $\ln\frac{P+t}{P}$ with slope $1/G_l$. On the real-corpus pairs this holds with $R^2\ge0.99$ from layer 3 on (Pythia-160m: $G_l=0.35, 0.74, 0.99, 1.13$ in layers $3, 6, 9, 11$, $R^2\ge0.992$; Pythia-1b in Table~\ref{tab:keff}), at least as well as the pure logarithmic form. In Pythia-1b the increments of $G_l$ per layer are about $0.65/l$, i.e.\ $G_l\approx0.65\ln l+\mathrm{const}$, as expected if the gain of a layer shrinks as the residual stream it writes into grows; these are effective parameters rather than estimates of the accumulated gain. Over our windows the two descriptions cannot be told apart.

\paragraph{Scale: Pythia-1b.} Table~\ref{tab:scale} repeats the real-corpus measurement on Pythia-1b (16 layers, width 2048) with the same 100 pairs per axis. At initialisation the exponents coincide with those of Pythia-160m at matched layers to two decimals (L1--L3: $0.73, 0.57, 0.50$ for both), the local exponent rises monotonically in every layer, and the fitted logarithmic powers are $k=1.33, 2.14, 2.78$ in layers 2, 3, 4, close to $l-1$ ($R^2\ge0.97$), with $k/\ln l\approx2.0$--$2.1$ beyond. After training, layer 1 is at $0.87$, layer 2 at $0.79$, and deep layers settle at $0.58$--$0.61$; paraphrase is again forgotten most slowly ($0.52$ in the last layer) and the unrelated axis again matches the one-word detail axis. Across a $6\times$ increase in parameters we see no qualitative change.

\paragraph{A second model family: Qwen2.5.} Qwen2.5 differs from Pythia in its normalisation (RMSNorm), its residual structure (sequential) and its attention (grouped-query); we measure it on the real-corpus pairs with $t_e\in[16,2048]$ (Table~\ref{tab:scale}). Trained Qwen2.5-0.5B gives $0.74$ in layer 1, $0.69$ in layers 2--3 and $0.54$--$0.62$ beyond; Qwen2.5-1.5B gives $0.95$ in layer 1, $0.81$ in layer 2 and $0.55$--$0.63$ beyond, the same profile as the trained Pythia models. At random initialisation $\alpha$ falls from $0.63$ (layer 1) to $0.18$ (layer 23), and the local exponent rises monotonically with $t$ in every layer. In the untrained model a one-character change of the prompt remains linearly decodable after 512 tokens; in the trained model it does not. Section~\ref{sec:results-qwen} discusses where Qwen departs from the linear relay picture.

\begin{table}[t]
\centering\small
\begin{tabular}{llccccccc}
\toprule
model & & L1 & L2 & L3 & & & & last \\
\midrule
160m & init    & 0.73 & 0.57 & 0.50 & L6 0.41 & L9 0.37 & & L11 0.36 \\
     & trained & 0.97 & 0.81 & 0.61 & L6 0.58 & L9 0.63 & & L11 0.66 \\
1b   & init    & 0.73 & 0.57 & 0.50 & L4 0.46 & L8 0.38 & L12 0.35 & L15 0.33 \\
     & trained & 0.87 & 0.79 & 0.71 & L4 0.69 & L8 0.61 & L12 0.61 & L15 0.58 \\
\midrule
Qwen-0.5B & init    & 0.63 & 0.45 & 0.37 & L6 0.28 & L12 0.22 & L18 0.20 & L23 0.18 \\
          & trained & 0.74 & 0.69 & 0.69 & L6 0.62 & L12 0.55 & L18 0.54 & L23 0.58 \\
Qwen-1.5B & trained & 0.95 & 0.81 & 0.61 & L7 0.63 & L14 0.62 & L21 0.56 & L27 0.55 \\
\bottomrule
\end{tabular}
\caption{$\alpha$ on real-corpus prompt pairs (mean over the four axes; per-axis 95\% intervals have half-width $\le0.06$), unbiased signal, windows labelled by their centre. Pythia-160m and Pythia-1b: $t_e\in[16,1024]$; Qwen2.5-0.5B and 1.5B: $t_e\in[16,2048]$. At initialisation the two Pythia sizes agree at matched layers.}
\label{tab:scale}
\end{table}

\paperfig{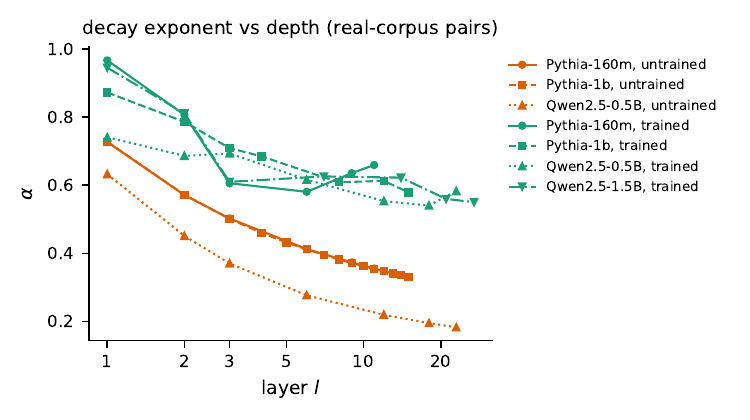}{0.5}{$\alpha$ against the layer index (mean over the four axes, real-corpus pairs). Untrained models (orange) depend on the layer index alone: the two Pythia sizes coincide; trained models (green) are flat beyond the first layers.}{fig:depth}

The \emph{unrelated} axis gives the sharpest test of the geometric picture: two completely different sentences are forgotten at the same rate as a one-word substitution (Pythia-160m, layer 11: $0.35$ vs.\ $0.37$ at initialisation, $0.70$ vs.\ $0.69$ after training). The size of the difference sets the amplitude of the offset, not its decay exponent.

\subsection{Signal versus mean squared offset}
\label{sec:results-ms}

In untrained Pythia the exponents of $\mathcal S$ and of the mean squared offset $\mathcal M$ coincide to within $0.01$ in every layer: the offset points in the same direction whatever the continuation. After training $\mathcal M$ decays more slowly in the deep layers (Pythia-160m: $\alpha_{\mathcal M}=0.52$--$0.57$ against $\alpha_{\mathcal S}=0.58$--$0.67$ in layers 6--11; Qwen2.5-0.5B: $0.52$--$0.58$ against $0.55$--$0.62$), and in untrained Qwen much more slowly ($0.10$--$0.21$ against $0.18$--$0.26$ in layers 6--23). Part of the decay of $\mathcal S$ in these models is therefore the decorrelation of the offset's direction across continuations, as anticipated in Remark~\ref{rem:jensen}; the norm of the offset itself decays more slowly.

\subsection{Where Qwen differs}
\label{sec:results-qwen}

Two features of the untrained Qwen model do not carry over from Pythia. First, the layer-1 local exponent does not follow the dilution law with a single $P_{\mathrm{eff}}$ (residuals up to $0.13$), and the fitted powers are $k=2.9, 4.5$ in layers 2, 3 ($k/\ln l\approx4$--$5$ beyond; $G_l=0.56$--$4.9$). The linear relay model gives exactly the dilution law in layer 1 and a power of exactly $1$ in layer 2 for any gains (Section~\ref{sec:theory-linear}), so the first blocks of untrained Qwen are not in the linear-relay regime, although the qualitative predictions (local exponent rising towards $1$, decreasing with depth) hold. This is not in conflict with Theorem~\ref{thm:nohair}, which bounds $\delta$ asymptotically and says nothing about its local slope at finite $t$. Attention in untrained Qwen is as uniform as in untrained Pythia (normalised entropy $0.990$--$0.993$, largest weight $0.006$--$0.007$ in layers 1--3, logit spread median $\approx2$), so non-uniform attention is not the reason. A difference that is present is the block structure: Pythia uses parallel residual connections, so its first MLP sees only the embedding, which is the same in both runs, and the layer-1 offset is exactly the attention write; Qwen's blocks are sequential, so the MLP of the first block already processes the attention write nonlinearly, and at initialisation that write dominates the residual (mid-block norm $0.033$, against about $0.02$ for embeddings drawn with standard deviation $0.02$). We take this as the likely reason the linear relay model fails in the first blocks of Qwen, without having tested it. Second, the noise across continuations grows slowly at initialisation ($\beta\approx-0.1$ to $-0.2$ in layers 2--23), so the decodability of the prompt decays as $t^{-2(\alpha-\beta)}$ rather than $t^{-2\alpha}$.

\paragraph{A late slowdown in trained Qwen.} With windows up to $2048$, the local exponent of the trained Qwen models decreases at late times. In Qwen2.5-0.5B this happens in every layer, down to $0.23$--$0.57$ in the last octave, including layer 1 ($0.24$), where only the direct path exists. In Qwen2.5-1.5B layers 1--2 keep decaying fast (last octave $2.2$ and $1.3$) while layers 3--27 slow down to $0.38$--$0.53$. In layers 14 and 21 of Qwen2.5-1.5B the signal stays below three times its sampling floor at all windows, so those two layers are not interpreted.

\emph{Three explanations excluded.} We first excluded a statistical explanation: with $n_c=8$ continuations the plug-in signal is biased upwards by $\operatorname{Var}_c\Delta/n_c$, which would flatten the curve late; with the unbiased estimator the slowdown is unchanged (Qwen2.5-0.5B, last octave: $0.24, 0.42, 0.43$ in layers 1--3). We then tested a prompt-specific attention sink: Lemma~\ref{lem:dilution} bounds the attention to the prompt by $Pe^{2B}/i$, which only bites once $i\gg e^{2B}$, and without a BOS token the sink is the first prompt token, which differs between the two prompts of most pairs. Prepending the same token (\texttt{<|endoftext|>}) to both prompts, which makes the sink shared, leaves the slowdown unchanged (last octave $0.24, 0.40, 0.44$ in layers 1--3). Finally we tested the structure of the continuation text: WikiText chunks span headings and document boundaries, which affect all prompts alike and are only probed at the largest windows. Replacing the continuations by WikiText sentences in random order, which removes headings, document boundaries and any structure beyond a sentence, leaves the slowdown unchanged as well (last octave $0.23, 0.49, 0.47$ in layers 1--3 and $0.18$--$0.57$ in the deep layers, against $0.24, 0.42, 0.43$ and $0.23$--$0.57$); the global exponents move by at most $0.06$ (slightly lower in the deep layers). Three explanations are thus excluded: estimation bias, a prompt-specific sink, and the long-range structure of the continuation.

\emph{A positional oscillation.} Since the slowdown appears in layer 1, where only the direct path exists, what remains is the dependence of attention on distance. We measured it directly: in the first block of trained Qwen2.5-0.5B, the attention that continuation queries pay to the prompt follows $1/t$ on average out to $t=2048$, with $(P+t)$ times the attention mass between $6$ and $11$ from $t=64$ on, but it oscillates by about $\pm30\%$ around that law from octave to octave (log-slopes $-0.94,-0.42,-1.25,-1.38,-0.64$ between consecutive windows from $t=64$). The local exponents of the layer-1 offset follow these oscillations: the steep octave $512$--$1024$ (attention slope $-1.38$) is where $\alpha_{\mathrm{loc}}$ exceeds $1$, and the shallower last octave ($-0.64$) is where it drops to $0.24$. The late slowdown in layer 1 is therefore an oscillation of the direct path around dilution, of positional origin, rather than a breakdown of dilution; its amplitude in the offset is larger than in the attention mass, so the value vectors contribute as well. In block 2, $(P+t)$ times the attention mass rises slowly with $t$ ($12$ to $17$ between $t=128$ and $2048$), a mild departure from $1/t$ at long range. The slowdown is absent from layers 1--2 of Qwen2.5-1.5B, whose attention we have not measured; tying the period of the oscillation to the rotary frequencies is left open. A plausible physical mechanism is phase cancellation in the rotary embeddings: at long distances, the dot product between different rotary frequency components can destructively interfere, causing the query to lose its original directional preference towards the prompt and oscillate around the \(1/t\) dilution law. We leave a rigorous proof of this interference to future work, but it aligns with the observation that the oscillation amplitude is larger in the offset than in the raw attention mass, as the value vectors also carry positional phase information.

\subsection{Continuation structure}
\label{sec:results-struct}

For trained networks on contiguous WikiText, $\alpha_{\mathrm{loc}}$ shows no systematic trend over $16$--$1920$ in the deep layers of Pythia (a power law over two decades), with a recurring bump at $128$--$256$ in both sizes. Repeating the measurement with shuffled-sentence continuations, which keep the vocabulary and the sentences of WikiText but remove all structure beyond a sentence, separates three effects (Table~\ref{tab:shuffled}).

\begin{table}[t]
\centering\small
\begin{tabular}{llcccccc}
\toprule
 & & \multicolumn{3}{c}{contiguous WikiText} & \multicolumn{3}{c}{shuffled sentences} \\
\cmidrule(lr){3-5}\cmidrule(lr){6-8}
model & layer & detail & para & unrel. & detail & para & unrel. \\
\midrule
160m, init    & L1  & 0.75 & 0.74 & 0.73 & 0.75 & 0.74 & 0.73 \\
              & L11 & 0.37 & 0.37 & 0.35 & 0.37 & 0.37 & 0.35 \\
160m, trained & L1  & 0.92 & 0.98 & 0.97 & 0.93 & 0.99 & 0.98 \\
              & L11 & 0.69 & 0.59 & 0.70 & 0.61 & 0.54 & 0.66 \\
1b, trained   & L1  & 0.90 & 0.86 & 0.88 & 0.90 & 0.86 & 0.88 \\
              & L15 & 0.60 & 0.52 & 0.62 & 0.48 & 0.41 & 0.53 \\
\bottomrule
\end{tabular}
\caption{$\alpha$ with contiguous WikiText continuations and with WikiText sentences in random order (real-corpus pairs, unbiased signal, centre-labelled windows).}
\label{tab:shuffled}
\end{table}

\begin{itemize}
  \item \textbf{Untrained networks do not notice.} For untrained Pythia-160m every exponent, every local exponent and every fitted $k$ and $G_l$ agrees with the contiguous-text values to within $0.01$: geometric dilution and relay at initialisation are independent of what the text says, as the theory requires.
  \item \textbf{The first layers do not notice either.} In trained models layers 1--4, where the direct path dominates, change by at most $0.04$. The bump of the layer-1 local exponent at $128$--$256$ in Pythia-160m survives ($1.62$ against $1.54$), so it is positional, not a property of the text.
  \item \textbf{Deep layers of trained models forget more slowly on unstructured text.} The global exponents drop by $0.02$--$0.11$ in layers 6--11 of Pythia-160m and by $0.09$--$0.18$ in layers 8--15 of Pythia-1b (by $\le0.06$ in Qwen2.5-0.5B). The shape of the local exponent changes in a model-dependent way: it rises from about $0.45$ to about $1.0$ across the window in Pythia-160m, and falls from about $0.5$ to about $0.3$ in Pythia-1b.
\end{itemize}

The pure power law of the trained deep layers is therefore a property of the model on coherent text. A natural reading is that a coherent continuation gives the deep layers relevant recent content to attend to, which displaces the prompt; without that structure the prompt competes for longer. We have not tested this mechanism. The ordering of the axes and the trained--untrained contrast are unaffected. In layers 1--2 of the trained models the local exponents fluctuate above $1$ in the last octaves, where the offset has decayed by more than three decades.

\section{Mechanism: direct path and relay}\label{sec:mechanism}
\subsection{The decomposition}\label{sec:mech-decomp}

The offset at a given layer is built from the attention writes of the blocks below it. To see where the prompt's imprint goes as it decays, we split each write into a \emph{direct} part, in which the query reads the prompt, and a \emph{relay} part, in which the query reads a continuation token that has already absorbed prompt information.

The write of attention block $k$ at query $q$ is $\sum_h\sum_ja^h_{qj}W_O^hv^h_j$, where $h$ runs over the heads of the block, $a^h_{qj}$ is the attention weight of head $h$ from query $q$ to key $j$, $W_O^h$ its output matrix and $v^h_j$ the value vector at key $j$. Splitting keys into prompt tokens (\emph{direct}) and continuation tokens (\emph{relay}) and differencing the two prompts under the same continuation gives an exact decomposition of the attention contribution to the offset; sinks shared by both prompts (a common first token) cancel.

\paperfig{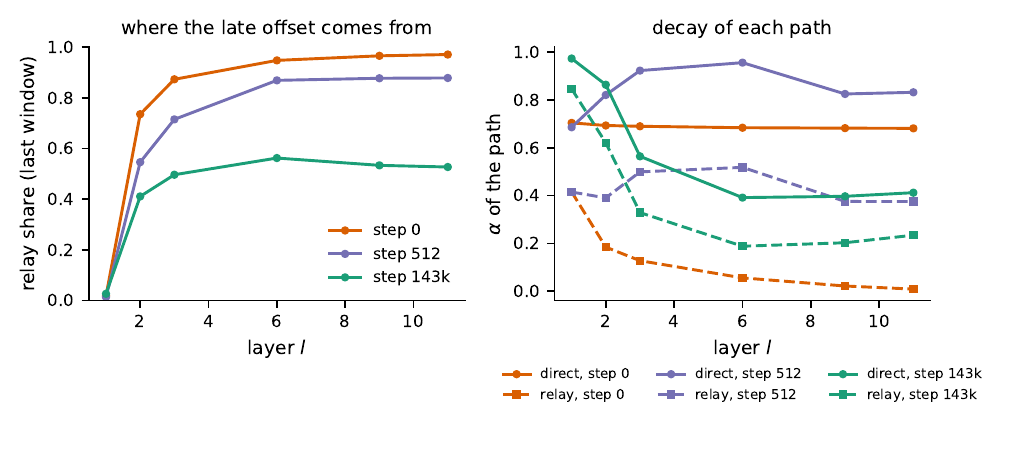}{1.0}{Path decomposition (Pythia-160m, real-corpus pairs). Left: share of the late offset carried by relay. Right: decay exponents of the direct (circles, solid) and relay (squares, dashed) paths. At initialisation relay carries almost all of the late offset and barely decays.}{fig:paths}

\begin{table}[t]
\centering\small
\begin{tabular}{llcccccc}
\toprule
step & & L1 & L2 & L3 & L6 & L9 & L11 \\
\midrule
0      & measured   & 0.71 & 0.56 & 0.49 & 0.40 & 0.36 & 0.35 \\
       & paths (both) & 0.70 & 0.53 & 0.46 & 0.38 & 0.34 & 0.32 \\
       & $\alpha_{\text{direct}}$ / $\alpha_{\text{relay}}$ & 0.70 / 0.42 & 0.69 / 0.18 & 0.69 / 0.13 & 0.68 / 0.06 & 0.68 / 0.02 & 0.68 / 0.01 \\
       & relay share & 0.02 & 0.73 & 0.87 & 0.95 & 0.97 & 0.97 \\
\midrule
512    & measured   & 0.70 & 0.74 & 0.79 & 0.77 & 0.66 & 0.67 \\
       & paths (both) & 0.69 & 0.73 & 0.79 & 0.73 & 0.60 & 0.61 \\
       & $\alpha_{\text{direct}}$ / $\alpha_{\text{relay}}$ & 0.69 / 0.42 & 0.82 / 0.39 & 0.92 / 0.50 & 0.96 / 0.52 & 0.83 / 0.38 & 0.83 / 0.38 \\
       & relay share & 0.01 & 0.55 & 0.71 & 0.87 & 0.88 & 0.88 \\
\midrule
143000 & measured   & 0.98 & 0.81 & 0.57 & 0.56 & 0.63 & 0.65 \\
       & paths (both) & 0.98 & 0.83 & 0.62 & 0.61 & 0.63 & 0.63 \\
       & $\alpha_{\text{direct}}$ / $\alpha_{\text{relay}}$ & 0.97 / 0.85 & 0.86 / 0.62 & 0.56 / 0.33 & 0.39 / 0.19 & 0.40 / 0.20 & 0.41 / 0.23 \\
       & relay share & 0.03 & 0.41 & 0.50 & 0.56 & 0.53 & 0.53 \\
\bottomrule
\end{tabular}
\caption{Path decomposition on real-corpus prompt pairs (Pythia-160m, 25 pairs per axis, four axes pooled, 8 WikiText continuations per pair, $t_e\in[16,1024]$, centre-labelled windows, unbiased signal for every component). Cumulative attention writes of blocks $0..l-1$ reproduce the measured $\alpha$ within $0.06$ at every layer and checkpoint. Relay share: relay part of the squared offset in the last window. Bootstrap half-widths of the measured $\alpha$ are $\le0.04$.}
\label{tab:paths}
\end{table}

Table~\ref{tab:paths} reports the decomposition at three checkpoints. Five patterns stand out.

\begin{itemize}
  \item \textbf{Layer 1 has no relay}, as it must: the continuation tokens are still identical after the first block, and the relay share is $0.02$--$0.03$ at every checkpoint.
  \item \textbf{At initialisation the late offset is 73--97\% relay}, rising with depth, and the relay component barely decays ($\alpha_{\text{relay}}$ from $0.18$ in layer 2 to $0.01$ in layer 11), while the direct path decays as a uniform head would ($\alpha_{\text{direct}}\approx0.68$--$0.70$ in every layer, the layer-1 value). The long tail of untrained networks is relay.
  \item \textbf{Training makes relay leak and steepens the direct path}: at step 512 $\alpha_{\text{relay}}\approx0.38$--$0.52$ and $\alpha_{\text{direct}}\approx0.82$--$0.96$ in layers 2--11, while relay still dominates the late offset ($0.55$--$0.88$).
  \item \textbf{After training the components no longer add up simply}: in layers 6--11 $\alpha_{\text{direct}}\approx0.39$--$0.41$ and $\alpha_{\text{relay}}\approx0.19$--$0.23$, yet their sum decays with $\alpha\approx0.61$--$0.63$. The two components have different decay rates and a cross term (they are anti-aligned at late times), so the exponent of the sum cannot be read off the exponents of the parts. The same happened on hand-written scenes from step $\sim$4000 on. Part of it is bookkeeping (softmax weights sum to one, so attention moved off the prompt lands on the continuation); in this regime only the sum is reliable.
  \item \textbf{Attention weights mislead.} Raw (or value-weighted) attention to the prompt predicts $\alpha$ only around steps 512--1000; before, it misses relay; after, sinks (token 0 and in-prompt delimiters) dominate the weights while carrying no prompt-specific information.
\end{itemize}

These patterns describe the mechanism at a fixed checkpoint. We now ask when the mechanism appears during training.

\subsection{When forgetting appears in training}\label{sec:mech-training}

\begin{table}[t]
\centering\small
\begin{tabular}{rcccccc}
\toprule
step & L1 & L2 & L3 & L6 & L9 & L11 \\
\midrule
0      & 0.73 & 0.57 & 0.50 & 0.41 & 0.37 & 0.36 \\
128    & 0.73 & 0.57 & 0.51 & 0.43 & 0.41 & 0.41 \\
256    & 0.72 & 0.62 & 0.62 & 0.57 & 0.52 & 0.52 \\
512    & 0.72 & 0.76 & 0.81 & 0.77 & 0.67 & 0.68 \\
1000   & 0.72 & 0.75 & 0.79 & 0.67 & 0.63 & 0.66 \\
4000   & 0.96 & 0.95 & 0.83 & 0.60 & 0.64 & 0.62 \\
16000  & 1.03 & 0.90 & 0.70 & 0.59 & 0.64 & 0.62 \\
143000 & 0.97 & 0.81 & 0.61 & 0.58 & 0.63 & 0.66 \\
\bottomrule
\end{tabular}
\caption{$\alpha$ across training, Pythia-160m, real-corpus prompt pairs (100 per axis, mean over the four axes, the same pairs at every checkpoint), $t_e\in[16,1024]$, centre-labelled windows, unbiased signal. Per-axis 95\% intervals have half-width $\le0.06$.}
\label{tab:alpha-train}
\end{table}

\paperfig{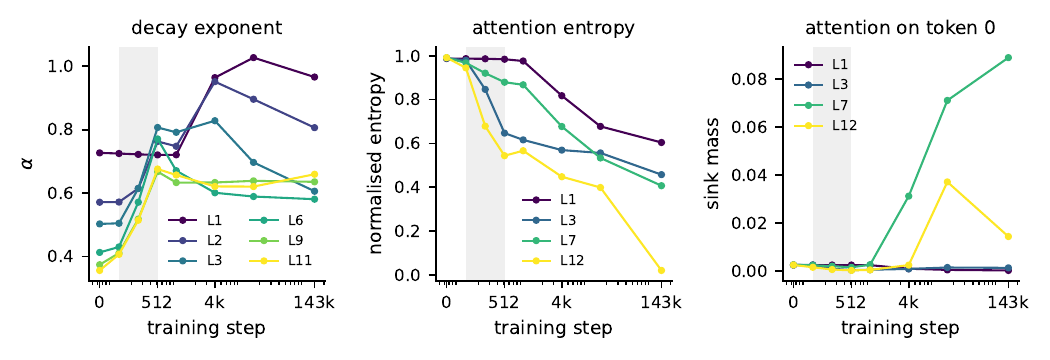}{1.0}{Training dynamics of Pythia-160m on the same 400 prompt pairs. Left: $\alpha$ per layer. Middle: normalised attention entropy. Right: attention mass on token 0. The shaded band marks steps 128--512, where forgetting sets in and attention beyond layer 1 leaves the uniform regime; the first layer and the sink change only after step 1000. Step 0 is drawn at the left edge.}{fig:training}

Table~\ref{tab:alpha-train} and Figure~\ref{fig:training} follow the same 400 prompt pairs through training. Three phases appear.

\begin{itemize}
  \item \textbf{Onset (steps 128--512, roughly 0.3--1B tokens).} Up to step 128 the exponents are those of the untrained network, and so is the attention: normalised entropy $\ge0.95$ in every layer measured. Between steps 128 and 512 every layer from 2 on jumps (layer 11: $0.41\to0.52\to0.68$); the path decomposition at step 512 shows that this is relay starting to leak (Table~\ref{tab:paths}). Attention leaves the uniform regime at the same time: the normalised entropy falls from $0.98$ to $0.85$ and $0.65$ in layer 3 and from $0.95$ to $0.68$ and $0.54$ in layer 12 between steps 128, 256 and 512, while layer 1 stays at $0.98$. Forgetting is strongest at step 512 and relaxes slightly afterwards in the deep layers.
  \item \textbf{First-layer locality (steps 1000--4000).} Layer 1 keeps the uniform-head value $0.72$--$0.73$ until step 1000 and reaches $0.96$ by step 4000: the first block becomes recency-biased only after forgetting in deeper layers is established, and its attention entropy falls only then ($0.98$ at step 1000, $0.82$ at step 4000). On the same 400 pairs, the attention mass on token 0 stays at or below $0.003$ in every layer measured (1, 2, 3, 7, 12) up to step 1000 and appears only afterwards ($0.031$ in layer 7 at step 4000, $0.071$ at 16000, $0.089$ at the end; $0.05$--$0.10$ in layers 5--11 of the final model).
  \item \textbf{Settling.} By the end of training deep layers sit at $0.58$--$0.66$, and the depth dependence of the untrained network ($0.73\to0.36$ from layer 1 to 11) has given way to a profile that is roughly flat beyond layer 3.
\end{itemize}

Forgetting therefore sets in between steps 128 and 512, earlier than the formation of attention sinks and earlier than the first layer becoming strongly recency-biased. The earliest change observed in training is that relay begins to leak, while layer 1, where only the direct path exists, stays in the untrained regime until step 1000.

\section{Causal test: compensating dilution}\label{sec:causal}
\subsection{The intervention}\label{sec:causal-setup}

The decomposition of Section~\ref{sec:mechanism} shows that the prompt's imprint decays through dilution, but it is a measurement, not an intervention. To test whether dilution is the cause of the decay, we cancel it.

We add $\gamma\log\frac{q+1}{P}$ to the logits from continuation queries $q$ to prompt tokens $1..P-1$, either uniformly ($\gamma=1$) or with per-head $\gamma_{l,h}$ fitted so that each head's attention to a typical prompt token stops decaying.
The boost exactly compensates the $1/t$ shrinkage of the prompt's share of attention under uniform attention.

\paperfig{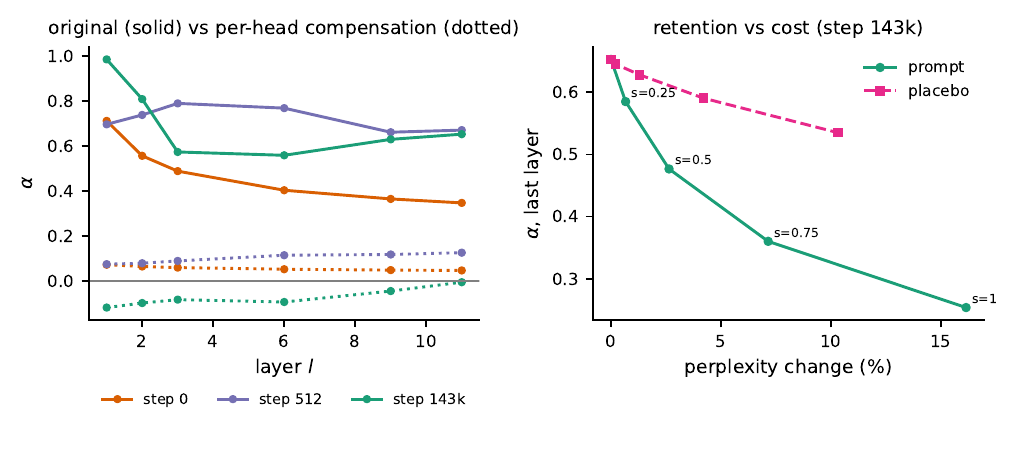}{1.0}{Left: $\alpha$ per layer before (solid) and after (dotted) per-head compensation of the dilution, at three checkpoints. Right: retention against cost at the end of training; the last-layer $\alpha$ against the perplexity change for a boost on the prompt and for an equally strong placebo boost.}{fig:causal}

\begin{table}[t]
\centering\small
\begin{tabular}{llcccccc}
\toprule
step & & L1 & L2 & L3 & L6 & L9 & L11 \\
\midrule
0      & original & 0.71 & 0.56 & 0.49 & 0.40 & 0.36 & 0.35 \\
       & uniform  & 0.10 & 0.09 & 0.08 & 0.07 & 0.06 & 0.06 \\
       & per-head & 0.07 & 0.06 & 0.06 & 0.05 & 0.05 & 0.05 \\
\midrule
512    & original & 0.70 & 0.74 & 0.79 & 0.77 & 0.66 & 0.67 \\
       & uniform  & 0.10 & 0.17 & 0.23 & 0.28 & 0.25 & 0.26 \\
       & per-head & 0.07 & 0.08 & 0.09 & 0.11 & 0.12 & 0.13 \\
\midrule
143000 & original & 0.98 & 0.81 & 0.57 & 0.56 & 0.63 & 0.65 \\
       & uniform  & 0.29 & 0.22 & 0.14 & 0.11 & 0.21 & 0.25 \\
       & per-head & $-0.12$ & $-0.10$ & $-0.08$ & $-0.09$ & $-0.04$ & $-0.01$ \\
\bottomrule
\end{tabular}
\caption{$\alpha$ under dilution compensation on real-corpus prompt pairs (same setting as Table~\ref{tab:paths}). Bootstrap half-widths $\le0.05$. Per-head calibration: residual median head slope $-0.07$ (step 0) and $-0.08$ (step 512), $-0.33$ (10--90\%: $[-1.16,-0.10]$) at step 143000.}
\label{tab:causal}
\end{table}

\subsection{What the intervention removes}\label{sec:causal-results}

Table~\ref{tab:causal} shows that compensating dilution removes forgetting at initialisation and at step 512; after training, uniform compensation leaves $\alpha=0.11$--$0.29$ and per-head compensation reverses the decay. Under uniform compensation the residual $\alpha$ matched the residual slope of prompt attention in a preliminary run on three hand-written scenes (steps 0, 256, 512: residual $\alpha=0.06$--$0.08$, $0.14$--$0.15$, $0.23$--$0.28$ against residual attention slopes $0.09$, $0.16$, $0.26$); per-head compensation brings $\alpha$ to $0.05$--$0.13$ at steps 0 and 512, and overshoots after training ($\alpha<0$: the prompt's influence grows), where the attention-based calibration does not converge, a sign that the attention a head pays to the prompt is not the same as the information it carries.

\subsection{What the test does and does not show}\label{sec:causal-caveat}

Per-head compensation raises perplexity by $140\%$ or more (Section~\ref{sec:application}), so the compensated network runs far from its training distribution. The test therefore shows that, in these networks, the forgetting we measure disappears once dilution is compensated: dilution is necessary for it. It does not show that the compensated network computes as the original one does, and it does not show that dilution is sufficient on its own; that it is sufficient is shown by the untrained networks and the linear relay model of Section~\ref{sec:theory-linear}, where uniform attention alone produces the decay. Within this caveat, no second forgetting mechanism is needed. Crucially, the placebo control (Table~\ref{tab:tradeoff}) provides a direct defense against the objection that the compensation simply destroys the model. While both the prompt boost and the placebo increase perplexity, only the prompt boost substantially flattens the decay of \(\alpha\). This asymmetry confirms that the retention is driven specifically by drawing attention back to the prompt, rather than by a global distortion of the next-token distribution.

In a preliminary run on the hand-written scenes, continuations generated by the model from the prompt gave the same $\alpha$ as human-written ones (within $0.1$; identical early in training), so the rate of forgetting is not inherited from the statistics of the model's own text; all real-corpus results above use human-written continuations.
\section{Exterior: what survives in the generated text}
\label{sec:exterior}

\paragraph{The question.} So far the continuation has been written by a human, so the exterior was the same for both prompts. In free generation the exterior is itself a product of the model, and the prompt can influence later tokens only through the text it has already caused. What properties of the prompt can still be read off from that text?

\paragraph{The probe.} The observer sees only text: a window of generated tokens ending at $t$ ($32$ tokens for Qwen, $8$ for Pythia) is turned into a hashed unigram histogram ($4096$ bins, square-root normalised), and a logistic-regression probe (inputs standardised, projected on $16$ principal components, $L_2$ penalty $C=0.1$) tells the two prompts of a pair apart under 5-fold stratified cross-validation. The numbers are accuracies averaged over three scenes, with $32$ freely generated trajectories per prompt (pure sampling at temperature $1$). Chance is $0.5$; the 95th percentile of a null obtained by splitting one prompt's trajectories at random is $0.58$--$0.63$. These runs predate the main pipeline and use three hand-written scenes (Chinese for Qwen, English with a French language variant for Pythia).

\paragraph{What survives.} Details, paraphrase and genre become undecodable from the text, while language remains decodable ($0.81$ for Qwen2.5-0.5B, $0.84$ for Pythia-160m at $t=512$). Language is not conserved: in Qwen, 4--11 of 32 trajectories per scene switch language, mostly towards Chinese, the model's dominant mode. The teacher-forced offset for the language axis still decays ($\alpha=0.54$--$0.65$ in layers 3--11 of trained Pythia-160m, against $0.61$--$0.69$ for a one-word detail; Table~\ref{tab:real}), so language survives through the trajectory itself, not through memory of the prompt.

\paragraph{Why language survives.} The two observations are complementary rather than contradictory. Corollary~\ref{cor:exterior} concerns the sensitivity of the next-token distribution to the prompt \emph{given the same history}; in free generation the history is itself a function of the prompt. Language therefore survives along an exterior channel: the prompt tilts the first tokens towards its language, and every later token is predicted from a history already written in that language, which keeps the trajectory there until it switches by chance. This is precisely the sense of the analogy: what persists is what the exterior field carries.

\section{Application: a dilution knob, and its costs}
\label{sec:application}

The analysis above says that the prompt's imprint fades because its share of attention shrinks as $1/t$. This suggests a simple intervention: boost the logits from the prompt by a factor that cancels the shrinkage, and see whether forgetting slows. We test this on Pythia-160m with a uniform boost $s\log\frac{q+1}{P}$ on the prompt, and compare it with a placebo of the same mean strength applied to the start of the continuation.

\begin{table}[t]
\centering\small
\begin{tabular}{lcccccc}
\toprule
 & \multicolumn{3}{c}{boost on the prompt} & \multicolumn{3}{c}{placebo boost} \\
\cmidrule(lr){2-4}\cmidrule(lr){5-7}
$s$ & $\alpha$ L6 & $\alpha$ L11 & $\Delta$PPL & $\alpha$ L6 & $\alpha$ L11 & $\Delta$PPL \\
\midrule
0    & 0.56 & 0.65 & --- & 0.56 & 0.65 & --- \\
0.25 & 0.45 & 0.58 & +0.7\% & 0.53 & 0.64 & +0.2\% \\
0.5  & 0.31 & 0.48 & +2.6\% & 0.48 & 0.63 & +1.3\% \\
0.75 & 0.20 & 0.36 & +7.1\% & 0.41 & 0.59 & +4.2\% \\
1    & 0.11 & 0.25 & +16.1\% & 0.34 & 0.53 & +10.3\% \\
\bottomrule
\end{tabular}
\caption{Trade-off of the uniform boost $s\log\frac{q+1}{P}$ on Pythia-160m (step 143000), real-corpus pairs (25 per axis, four axes pooled; centre-labelled windows, unbiased signal; $\alpha$ intervals $\le\pm0.06$). $\Delta$PPL: perplexity change on WikiText continuations of 40 real-corpus prompts. The placebo boosts a same-length span at the start of the continuation.}
\label{tab:tradeoff}
\end{table}

\subsection{The trade-off}
\label{sec:app-tradeoff}

\begin{itemize}
  \item A mild boost is cheap: $s=0.5$ lowers $\alpha$ by $0.17$--$0.25$ for $+2.6\%$ perplexity. The placebo also lowers $\alpha$, because the boosted continuation tokens follow the prompt directly and already carry its imprint through relay. At matched retention the prompt boost is much cheaper than the placebo ($\alpha_{\text{L11}}\approx0.5$: $+2.6\%$ at $s=0.5$ vs.\ $+10.3\%$ at $s=1$). At matched strength it is costlier on these pairs ($+16.1\%$ vs.\ $+10.3\%$ at $s=1$), whereas in a preliminary run on three hand-written scenes, whose continuations were generated from the prompt, the two cost about the same ($s=0.5$: $+2.7\%$ vs.\ $+2.2\%$ on human-written text, $+1.3\%$ vs.\ $+1.7\%$ on the model's own text; $s=1$: $+14.0\%$ vs.\ $+14.8\%$ and $+8.3\%$ vs.\ $+10.8\%$). The cost of remembering the prompt thus depends on its relevance to the continuation: here the WikiText continuations are unrelated to the prompt, and attending to it more hurts prediction.
  \item Per-head calibrated compensation is too strong to be practical (+140\% to +376\% perplexity).
\end{itemize}

\subsection{Instruction adherence and its side effects}
\label{sec:app-adherence}

\begin{itemize}
  \item Qwen2.5-0.5B-Instruct, system prompt ``answer only in French'', 2048 generated tokens: share of samples fully in French $31\%\to94\%$ at $s=0.5$. Qwen2.5-1.5B-Instruct: $84\%\to94$--$100\%$ for all boosts; a constant boost with the same mean bias does as well, with less echo.
  \item \textbf{Multi-turn conversation} (Qwen2.5-1.5B-Instruct, 8 turns, English user turns asking for the next chapter, natural reply endings, 32 conversations per condition). Without boost the French share drops from $0.90$ to $0.81$--$0.87$ after the first turn and 78\% of conversations stay in French throughout. With $\log t$ boosts of strength $0.1$ and $0.25$: last-three-turn French share $0.90$ and $0.96$, conversations fully in French 84\% and 91\%. Constant boosts with the same mean bias: $0.94$ and $0.94$, 84\% and 88\%. Echo in late turns stays at the baseline level (the excess is concentrated in the first reply, more so for the constant boost: $12.9$ vs.\ $8.6$ leak words per 1000 at the higher strength); repetition does not increase; replies are about 10\% shorter. With 32 conversations per condition these differences are trends rather than significant effects, and the $\log t$ shape shows no clear advantage over a matched constant: the benefit comes from extra attention to the system prompt.
  \item \textbf{Instruction echo.} With forced-length single-turn generation, the share of outputs whose second half talks about the instruction rises from 19\% to 56--69\% (1.5B); much of this text is filler written after the story has ended.
\end{itemize}

Across all three tests, the pattern is the same: what helps is extra attention to the instruction, not the shape of the correction.

\section{Discussion and conclusion}
\label{sec:discussion}

A transformer's memory of its prompt, measured as the prompt-induced offset of the hidden state under a shared continuation, decays approximately as a power law whose exponent we can measure, bound and manipulate. The bound follows from geometric dilution of attention and the number of relay rounds through depth; it is attained asymptotically in the uniform-attention regime, whose exact solution with one gain per layer accounts for the slow, depth-dependent forgetting of untrained networks; trained networks suppress relay by about one billion training tokens, before attention sinks appear, and come close to pure dilution. Compensating the dilution removes or reverses forgetting, so dilution is necessary for it in these networks; that it is also sufficient is shown by the untrained networks and the linear relay model, where uniform attention alone produces the decay. Two negative results matter as much: attention weights are not a reliable measurement of the effect, and a $\log t$-shaped compensation is no better than a constant boost for keeping an instruction alive; what helps is extra attention to the instruction, not the shape of the correction.

These findings come with limitations. The constant of Theorem~\ref{thm:nohair} can grow exponentially with depth and the bound is informative only for $t\gg Pe^{2B}$, so the theorem constrains the asymptotic shape of the decay rather than the exponents we fit; the non-degeneracy assumption is verified empirically rather than derived, with a small margin in the first block of Qwen. How the exponents of trained networks relate to the theorem is therefore an open question: for trained models the bound is uninformative within our windows, and the solvable model describes the untrained ones.), and in trained Qwen the late slowdown is an oscillation of layer-1 attention around $1/t$ whose period, and its relation to the rotary frequencies, remain open. The single-turn behavioural test forces a fixed generation length, so its late text partly consists of filler written after the story has ended; the multi-turn test avoids this but has only 32 conversations per condition. Exponents are fitted over at most seven octaves, which does not allow the logarithmic and small-gain descriptions of the untrained curves to be told apart.

Several questions are left open. The most direct is scale: whether the exponents we fit stay in the same range at larger models, and whether the depth dependence of untrained networks survives. A second is the period of the layer-1 oscillation in Qwen, which we have not tied to the rotary frequencies. A third is whether the picture changes under instruction tuning: all our trained models are pretrained only, and the path decomposition across checkpoints suggests that the training process itself is what suppresses relay, so a model trained on dialogue may forget its system prompt at a different rate. Finally, the toolkit we release makes it easy to measure $\alpha$ on any Hugging Face model, and we hope the community will extend the measurements beyond the range we could afford.

Conditional on the generated text the prompt is forgotten; what survives, such as the language of the conversation, is carried by the text itself.

\bibliographystyle{plainnat}
\bibliography{refs}

\appendix
\section{Proofs}
\label{app:proofs}

This appendix gives the steps that Section~\ref{sec:theory} compresses: a perturbation lemma for softmax, the bookkeeping of the constants over heads, the alignment of positions when the two prompts differ in length, a complete proof of Lemma~\ref{lem:recursion}, and a quantitative version of Corollary~\ref{cor:smallgain} with explicit error terms.

\subsection{Softmax along a path}

\begin{lemma}\label{lem:softmax}
For $s,s'\in\mathbb R^n$ let $s_\theta=s+\theta(s'-s)$ and $a(\cdot)$ the softmax. Then
\[
\|a(s')-a(s)\|_1\;\le\;2\sup_{\theta\in[0,1]}\sum_{j}a_j(s_\theta)\,|s'_j-s_j| .
\]
If moreover $|s_j|,|s'_j|\le B$ for all $j$, then $a_j(s_\theta)\le e^{2B}/n$ for all $j$ and $\theta$.
\end{lemma}
\begin{proof}
With $\Delta=s'-s$ and $a=a(s_\theta)$, $\frac{d}{d\theta}a(s_\theta)=a\odot\big(\Delta-\langle a,\Delta\rangle\mathbf 1\big)$ (with $\odot$ the elementwise product), whose $\ell_1$ norm is at most $\sum_ja_j|\Delta_j|+|\langle a,\Delta\rangle|\sum_ja_j\le2\sum_ja_j|\Delta_j|$. Integrating over $\theta\in[0,1]$ gives the first claim. Each coordinate of $s_\theta$ is a convex combination of $s_j$ and $s'_j$, hence lies in $[-B,B]$, and $a_j(s_\theta)\le e^{B}/(ne^{-B})$.
\end{proof}

\subsection{Constants of a multi-head block}

An attention block with $H$ heads writes $\sum_{h=1}^{H}W_O^{(h)}\sum_ja^{(h)}_{ij}W_V^{(h)}\mathrm{LN}(h_j)$, where $H$ is the number of heads, $a^{(h)}_{ij}$ are the attention weights of head $h$, $W_V^{(h)}$ its value matrix, and $W_O^{(h)}$ the block of columns of $W_O$ that reads head $h$. For each head let
\[
B_h=\frac{d\,\|W_Q^{(h)}\|\,\|W_K^{(h)}\|}{\sqrt{d_h}}+\beta_h,\quad
V_h=\sqrt d\,\|W_O^{(h)}W_V^{(h)}\|,\quad
\rho_h=\frac{2\sqrt d}{r}\|W_O^{(h)}W_V^{(h)}\|,\quad
\lambda_h=\frac{2d\,\|W_Q^{(h)}\|\,\|W_K^{(h)}\|}{r\sqrt{d_h}},
\]
with $\beta_h$ the contribution of query and key biases, if any. The constants of Section~\ref{sec:theory} are to be read as
\[
B=\max_hB_h,\qquad V=\sum_hV_h,\qquad \rho=\sum_h\rho_h,\qquad \lambda V:=\sum_h\lambda_hV_h ,
\]
so that every bound proved for one head with $(B_h,V_h,\rho_h,\lambda_h)$ sums over heads into the same bound with $(B,V,\rho,\lambda V)$; in particular $K_1,K_2,K_3$ of Lemma~\ref{lem:recursion} carry no hidden factor $H$. Gains of the normalisation layers are absorbed into the adjacent weight matrices. Lemma~\ref{lem:consts} then holds head by head with these constants, by the same arguments.

\subsection{Positions}

\begin{lemma}\label{lem:positions}
With rotary position encodings, the logit between a query at continuation distance $i$ and a key at continuation distance $j\le i$ depends on the positions only through $i-j$; in particular it is the same function of $(h_{l-1}(i),h_{l-1}(j))$ in both runs, whatever the prompt lengths $P$ and $P'$.
\end{lemma}
\begin{proof}
Rotary encodings give $s=\langle R_{p}q,R_{p'}k\rangle/\sqrt{d_h}=\langle q,R_{p'-p}k\rangle/\sqrt{d_h}$ because $R_p$ is orthogonal and $R_p^\top R_{p'}=R_{p'-p}$, where $q$ and $k$ are the query and key vectors, $p,p'$ their absolute positions and $R_p$ the rotary matrix at position $p$. A continuation token at distance $j$ sits at absolute position $P+j$ in one run and $P'+j$ in the other, so $p'-p=j-i$ in both. Positions enter nowhere else in the block.
\end{proof}

With absolute position embeddings the continuation tokens would carry different embeddings in the two runs; we then take $P=P'$, which restores $\delta_0\equiv0$ at continuation positions.

\paragraph{Scope.} The proof covers multi-head attention with full or partial rotary encodings or with no positional term in the logits, grouped-query attention (key and value matrices shared across heads only change the per-head constants), query and key biases, LayerNorm and RMSNorm, and sequential or parallel blocks; absolute position embeddings are covered for $P=P'$. It does not cover additive position biases that grow with distance, such as ALiBi: they make the logits unbounded, so Lemma~\ref{lem:consts}(a) fails, even though they push towards more forgetting rather than less; nor logit scalings that grow with context length, such as Scalable-Softmax \citep{nakanishi2025ssmax}, which also make the logits unbounded and, unlike ALiBi, can push towards less forgetting; nor sliding-window attention, for which the number of visible keys stops growing and the dilution lemma no longer applies as stated; nor non-softmax attention.

\subsection{Proof of Lemma~\ref{lem:recursion}}

Fix layer $l$, a continuation query at distance $i$ and a head $h$; drop the head index. Write $u_j=W_OW_V\mathrm{LN}(h_{l-1}(j))$ and $u'_j$ for the second run, so $\|u_j\|,\|u'_j\|\le V$, and $\|u_j-u'_j\|\le\rho\,\delta_{l-1}(j)$ at continuation keys. Let $\pi=\sum_{j\in\text{prompt}}a_{ij}$, $\hat a_{ij}=a_{ij}/(1-\pi)$ for the $i$ continuation keys, $\bar a_{\mathrm P}=\pi^{-1}\sum_{j\in\text{prompt}}a_{ij}u_j$ and $\bar a_{\mathrm C}=\sum_{j\le i}\hat a_{ij}u_j$, and similarly with primes. The head output is $a=\pi\bar a_{\mathrm P}+(1-\pi)\bar a_{\mathrm C}$, and
\[
a-a'=(1-\pi)(\bar a_{\mathrm C}-\bar a'_{\mathrm C})+(\pi'-\pi)\bar a'_{\mathrm C}+\pi\bar a_{\mathrm P}-\pi'\bar a'_{\mathrm P}.
\]
\emph{Prompt terms.} The query sees $n=P+i\ge i$ keys in one run and $P'+i\ge i$ in the other; by Lemma~\ref{lem:dilution}, $\pi,\pi'\le Pe^{2B}/i$, so $|\pi'-\pi|\le Pe^{2B}/i$ and, since $\|\bar a_{\mathrm P}\|,\|\bar a'_{\mathrm C}\|\le V$, the last three terms have norm at most $3Ve^{2B}P/i$.

\emph{Continuation term.} $\hat a_{i\cdot}$ and $\hat a'_{i\cdot}$ are softmaxes over the same $i$ keys of the logits $s_{ij}$ and $s'_{ij}$, $j\le i$, which lie in $[-B,B]$; by Lemma~\ref{lem:positions} they differ only through the hidden states. Then
\[
\bar a_{\mathrm C}-\bar a'_{\mathrm C}=\sum_{j\le i}\hat a_{ij}(u_j-u'_j)+\sum_{j\le i}(\hat a_{ij}-\hat a'_{ij})u'_j .
\]
By Lemma~\ref{lem:softmax}, $\hat a_{ij}\le e^{2B}/i$, so the first sum is at most $\rho e^{2B}i^{-1}\sum_{j\le i}\delta_{l-1}(j)$. For the second, Lemma~\ref{lem:softmax} and $|s_{ij}-s'_{ij}|\le\lambda\big(\delta_{l-1}(i)+\delta_{l-1}(j)\big)$ (Lemma~\ref{lem:consts}(c)) give
\[
\Big\|\sum_j(\hat a_{ij}-\hat a'_{ij})u'_j\Big\|\le V\|\hat a_{i\cdot}-\hat a'_{i\cdot}\|_1\le2\lambda V\Big(\delta_{l-1}(i)+\frac{e^{2B}}{i}\sum_{j\le i}\delta_{l-1}(j)\Big),
\]
using $\sum_j\hat a_{ij}(s_\theta)=1$ for the first term and $\hat a_{ij}(s_\theta)\le e^{2B}/i$ for the second. With $1-\pi\le1$ and the sum over heads of the previous subsection,
\[
\|a-a'\|\le2\lambda V\,\delta_{l-1}(i)+e^{2B}(\rho+2\lambda V)\,\frac1i\sum_{j\le i}\delta_{l-1}(j)+3Ve^{2B}\frac Pi .
\]
\emph{Residual and MLP.} In a sequential block, $h_{\mathrm{mid}}=h_{l-1}+a$ and $h_l=h_{\mathrm{mid}}+M(\mathrm{LN}(h_{\mathrm{mid}}))$, with $M\circ\mathrm{LN}$ $\mu$-Lipschitz on non-degenerate inputs (Assumption~\ref{ass:nondeg}), so $\delta_l(i)\le(1+\mu)\big(\delta_{l-1}(i)+\|a-a'\|\big)$. In a parallel block, $h_l=h_{l-1}+a+M(\mathrm{LN}(h_{l-1}))$, so $\delta_l(i)\le(1+\mu)\delta_{l-1}(i)+\|a-a'\|$, which is smaller. Inserting the bound on $\|a-a'\|$ gives \eqref{eq:recursion} with $K_1=(1+\mu)(1+2\lambda V)$, $K_2=(1+\mu)e^{2B}(\rho+2\lambda V)$ and $K_3=3(1+\mu)e^{2B}V$. \qed

\subsection{A quantitative form of Corollary~\ref{cor:smallgain}}

Let $T_k(G,v)=\dfrac{G^kv^{k-1}}{k!\,(k-1)!}$, so that $\varphi(G,u)=\sum_{k\ge1}T_k(G,u)$ in \eqref{eq:bessel0}.

\begin{lemma}\label{lem:hbounds}
With $x_j=1/(P+j)$, $u=\ln\frac{P+i}{P}$ and $k\ge0$, and writing $(\cdot)_+=\max(\cdot,0)$ for the positive part,
\[
\frac{\big((u-1/P)_+\big)^{k}}{k!}\;\le\;h_k(x_1,\dots,x_i)\;\le\;\frac{\big(u+\frac{k-1}{2P}\big)^{k}}{k!}.
\]
\end{lemma}
\begin{proof}
The lower bound combines $h_k\ge p_1^k/k!$ and $p_1>u-1/P$ from the proof of Proposition~\ref{prop:gains}. For the upper bound, $\sum_kh_kz^k=\prod_j(1-x_jz)^{-1}=\exp\big(\sum_{m\ge1}p_mz^m/m\big)$ with $p_m=\sum_jx_j^m\le x_{\max}^{m-1}p_1\le p_1/P^{m-1}$ (where $x_{\max}=\max_jx_j$). Exponentials of series with non-negative coefficients preserve coefficientwise inequalities, so $h_k$ is at most the coefficient of $z^k$ in $\exp\big(Pp_1\sum_m(z/P)^m/m\big)=(1-z/P)^{-Pp_1}$, namely $\prod_{r=0}^{k-1}(p_1+r/P)/k!$. By the AM--GM inequality this is at most $(p_1+\frac{k-1}{2P})^k/k!$, and $p_1\le u$.
\end{proof}

\begin{proposition}\label{prop:sandwich}
In the linear relay model, with $G=G_l$, $Q=Q_l$ and $u=\ln\frac{P+i}{P}$, for all $l,i\ge1$,
\[
\sum_{k=1}^{l}\Big(1-\frac{Q\,k(k-1)}{2G^2}\Big)_+\,T_k\big(G,(u-\tfrac1P)_+\big)\;\le\;\frac{P+i}{P}\,\delta_l(i)\;\le\;\sum_{k=1}^{l}T_k\Big(G,u+\frac{k-2}{2P}\Big).
\]
\end{proposition}
\begin{proof}
Insert into \eqref{eq:gains} the bounds $\big(1-\frac{Qk(k-1)}{2G^2}\big)_+\frac{G^k}{k!}\le e_k\le\frac{G^k}{k!}$, which restate $0\le G^k/k!-e_k\le\frac Q2G^{k-2}/(k-2)!$ from the proof of Corollary~\ref{cor:smallgain}, and the bounds of Lemma~\ref{lem:hbounds} on $h_{k-1}$. All terms are non-negative.
\end{proof}

We checked both inequalities numerically against the exact recursion \eqref{eq:linrelay} for 300 random choices of depth ($l\le24$), prompt length ($5\le P\le60$) and gains ($0.002\le g_m\le2$), at $i\in\{1,3,10,50,300,2000\}$: no violation in $1800$ checks.

\begin{corollary}\label{cor:smallgain-quant}
Write $w_k=T_k(G,u)/\varphi(G,u)$ for the normalised terms of the series, $m_r=\sum_kk^rw_k$, and $\tau=\sum_{k>l}w_k$ for the weight beyond depth $l$. Then
\[
-\Big(\frac{Q\,m_2}{2G^2}+\frac{m_1}{Pu}+\tau\Big)\;\le\;\frac{(P+i)\,\delta_l(i)}{P\,\varphi(G,u)}-1\;\le\;\sum_kw_k\Big(e^{\frac{(k-1)(k-2)}{2Pu}}-1\Big),
\qquad
\tau\le\frac{T_{l+1}(G,u)}{\varphi(G,u)}\cdot\frac{1}{1-\frac{Gu}{(l+1)(l+2)}}
\]
whenever $Gu<(l+1)(l+2)$.
\end{corollary}
\begin{proof}
For the lower bound use $(u-1/P)^{k-1}\ge u^{k-1}\big(1-\frac{k-1}{Pu}\big)$ and $(1-a)(1-b)\ge1-a-b$ for $a,b\ge0$ in Proposition~\ref{prop:sandwich}, drop the factor $(\cdot)_+$ in favour of the linear bound, and add the missing terms $k>l$ as $-\tau$. For the upper bound use $\big(1+\frac{k-2}{2Pu}\big)^{k-1}\le e^{(k-1)(k-2)/(2Pu)}$ and extend the sum to all $k$. For $\tau$, the ratio of consecutive terms is $T_{k+1}/T_k=Gu/\big(k(k+1)\big)\le Gu/\big((l+1)(l+2)\big)$ for $k\ge l+1$; sum the geometric series.
\end{proof}

\paragraph{Reading the bound.} The weights $w_k\propto z^{k-1}/\big(k!(k-1)!\big)$, $z=Gu$, are those of the series of $I_1(2\sqrt z)$; for $z\gg1$ they concentrate around $k\approx\sqrt z$ (the mean of $k-1$ is $\zeta(z)=\sqrt z-\frac34+O(z^{-1/2})$, from \eqref{eq:bessel}), so $m_1\approx\sqrt z$ and $m_2\approx z$. The relative error of the small-gain form is then of order
\[
\frac{Q_lu}{2G_l}\;+\;\frac1P\sqrt{\frac{G_l}{u}}\;+\;\frac{G_l}{2P}\;+\;\tau ,
\]
which are the conditions stated in Corollary~\ref{cor:smallgain}: $Q_lu\ll G_l$ (many small gains; $Q_l\ll G_l^2$ is not enough, since $Q_lu/G_l=(Q_l/G_l^2)\,G_lu$), prompts long compared with $G_l$, and depth large compared with $\sqrt{G_lu}$. For $z\lesssim1$, as in the first layers of our measurements, $w_1$ dominates and all corrections are small, but the asymptotic expansion \eqref{eq:bessel} of the local exponent is not valid there; this is why the fitted $G_l$ of Section~\ref{sec:theory-linear} are effective parameters. Statements (a) and (b) of Corollary~\ref{cor:smallgain} are properties of $\varphi$; a bound on $\delta$ such as the one above does not by itself bound its logarithmic derivative, so the local exponents of the model are covered by (a) and (b) only to the extent that $\delta$ is well approximated by $\varphi$ uniformly over an octave.

\section{Non-degeneracy of pre-norm activations}
\label{app:nondeg}
Table~\ref{tab:nondeg} reports $\|\Pi h\|/\sqrt d$ at the input of every block and of the final norm, over 80 sequences of length $T=1024$ (40 real-corpus prompts $\times$ 2 WikiText continuations). The minimum is never close to zero on these sequences; this supports Assumption~\ref{ass:nondeg} empirically but does not prove it for other continuations or longer contexts. The embeddings (block 0 input) are small but are not used by Theorem~\ref{thm:nohair} (Assumption~\ref{ass:nondeg}). For the trained Pythia-1b the minimum over blocks 1--15 is $0.49$ (median $0.57$--$2.81$) and $0.97$ at the final norm, again a wide margin. Qwen2.5-0.5B uses RMSNorm ($\Pi=I$); its minimum over block inputs 1--23 is $0.105$ (median $0.17$--$2.56$) and $1.30$ at the final norm. Because its blocks are sequential, the post-attention norms enter Assumption~\ref{ass:nondeg} as well; their minimum is $0.012$, in the first block, so $r/\sqrt d=0.012$ for this model.

\begin{table}[h]
\centering\small
\begin{tabular}{lcccccc}
\toprule
 & \multicolumn{3}{c}{init (step 0)} & \multicolumn{3}{c}{trained (step 143000)}\\
\cmidrule(lr){2-4}\cmidrule(lr){5-7}
input of & min & 0.01\% & median & min & 0.01\% & median \\
\midrule
block 0 (embeddings) & 0.021 & 0.021 & 0.023 & 0.018 & 0.018 & 0.028 \\
block 1  & 0.112 & 0.112 & 0.127 & 0.236 & 0.241 & 0.325 \\
block 2  & 0.162 & 0.162 & 0.184 & 0.279 & 0.285 & 0.402 \\
block 3  & 0.203 & 0.208 & 0.229 & 0.320 & 0.331 & 0.477 \\
block 6  & 0.305 & 0.307 & 0.335 & 0.428 & 0.436 & 0.616 \\
block 9  & 0.377 & 0.378 & 0.414 & 0.519 & 0.529 & 0.883 \\
block 11 & 0.416 & 0.418 & 0.460 & 0.605 & 0.607 & 1.114 \\
final norm & 0.433 & 0.437 & 0.479 & 0.811 & 0.938 & 4.319 \\
\bottomrule
\end{tabular}
\caption{$\|\Pi h\|/\sqrt d$ for Pythia-160m. The theorem uses blocks $\ge1$ and the final norm: $r/\sqrt d=0.112$ (init) and $0.236$ (trained).}
\label{tab:nondeg}
\end{table}

\section{Prompt sets}
\label{app:prompts}
The twenty hand-written scenes. Each prompt is followed by the same sixteen WikiText-103 validation passages. Real-corpus pairs are drawn as described in Section~\ref{sec:measure-data}; the exact pairs are released with the toolkit.

{\small\begin{itemize}\setlength\itemsep{0.4em}
\item[1.] \begin{tabular}[t]{@{}ll@{}}\emph{base} & \parbox[t]{0.78\linewidth}{It was raining in Paris today, and I stood at the metro entrance,}\\\emph{detail} & \parbox[t]{0.78\linewidth}{It was snowing in Paris today, and I stood at the metro entrance,}\\\emph{para} & \parbox[t]{0.78\linewidth}{Rain was falling over Paris today; I stopped at the door of the metro station,}\\\emph{lang} & \parbox[t]{0.78\linewidth}{Il pleuvait à Paris aujourd'hui, et je me tenais à l'entrée du métro,}\\\emph{genre} & \parbox[t]{0.78\linewidth}{PARIS (Reuters) - Heavy rain hit the French capital on Tuesday, and at one metro entrance}\\\end{tabular}
\item[2.] \begin{tabular}[t]{@{}ll@{}}\emph{base} & \parbox[t]{0.78\linewidth}{In my grandmother's kitchen, rain ran down the window as she}\\\emph{detail} & \parbox[t]{0.78\linewidth}{In my grandmother's kitchen, snow piled up on the windowsill as she}\\\emph{para} & \parbox[t]{0.78\linewidth}{Rain streaked the glass of my grandmother's kitchen window while she}\\\emph{lang} & \parbox[t]{0.78\linewidth}{Dans la cuisine de ma grand-mère, la pluie coulait sur la vitre pendant qu'elle}\\\emph{genre} & \parbox[t]{0.78\linewidth}{LONDON (AP) - Days of heavy rain left several homes with leaking kitchen windows, and one elderly resident}\\\end{tabular}
\item[3.] \begin{tabular}[t]{@{}ll@{}}\emph{base} & \parbox[t]{0.78\linewidth}{It was raining that day. He waited at the train station for a long time, and finally}\\\emph{detail} & \parbox[t]{0.78\linewidth}{It was snowing that day. He waited at the train station for a long time, and finally}\\\emph{para} & \parbox[t]{0.78\linewidth}{The rain never stopped. He spent hours waiting at the station until, at last,}\\\emph{lang} & \parbox[t]{0.78\linewidth}{Il pleuvait ce jour-là. Il a attendu longtemps à la gare, et finalement}\\\emph{genre} & \parbox[t]{0.78\linewidth}{BERLIN (Reuters) - Rain delayed dozens of trains on Monday, and one passenger who had waited for hours}\\\end{tabular}
\item[4.] \begin{tabular}[t]{@{}ll@{}}\emph{base} & \parbox[t]{0.78\linewidth}{The old fisherman pulled his red boat onto the beach and}\\\emph{detail} & \parbox[t]{0.78\linewidth}{The old fisherman pulled his blue boat onto the beach and}\\\emph{para} & \parbox[t]{0.78\linewidth}{Dragging his red boat up onto the sand, the old fisherman}\\\emph{lang} & \parbox[t]{0.78\linewidth}{Le vieux pêcheur tira son bateau rouge sur la plage et}\\\emph{genre} & \parbox[t]{0.78\linewidth}{LISBON (AP) - An elderly fisherman hauled his boat ashore on Sunday as officials warned that}\\\end{tabular}
\item[5.] \begin{tabular}[t]{@{}ll@{}}\emph{base} & \parbox[t]{0.78\linewidth}{My sister opened the letter from the university, read the first line, and}\\\emph{detail} & \parbox[t]{0.78\linewidth}{My brother opened the letter from the university, read the first line, and}\\\emph{para} & \parbox[t]{0.78\linewidth}{After tearing open the university's letter, my sister glanced at the first sentence and}\\\emph{lang} & \parbox[t]{0.78\linewidth}{Ma sœur ouvrit la lettre de l'université, lut la première ligne, et}\\\emph{genre} & \parbox[t]{0.78\linewidth}{BOSTON (Reuters) - Thousands of students received admission letters this week, and one applicant}\\\end{tabular}
\item[6.] \begin{tabular}[t]{@{}ll@{}}\emph{base} & \parbox[t]{0.78\linewidth}{The cat jumped onto the piano in the middle of the night, and}\\\emph{detail} & \parbox[t]{0.78\linewidth}{The dog jumped onto the piano in the middle of the night, and}\\\emph{para} & \parbox[t]{0.78\linewidth}{Late at night the cat leapt up onto the piano keys, and}\\\emph{lang} & \parbox[t]{0.78\linewidth}{Le chat sauta sur le piano au milieu de la nuit, et}\\\emph{genre} & \parbox[t]{0.78\linewidth}{VIENNA (AP) - A cat triggered a late-night alarm at a concert hall on Friday after it}\\\end{tabular}
\item[7.] \begin{tabular}[t]{@{}ll@{}}\emph{base} & \parbox[t]{0.78\linewidth}{We drove north for six hours until the road ended at a frozen lake, where}\\\emph{detail} & \parbox[t]{0.78\linewidth}{We drove north for nine hours until the road ended at a frozen lake, where}\\\emph{para} & \parbox[t]{0.78\linewidth}{After six hours heading north, the road stopped at the edge of a frozen lake, and there}\\\emph{lang} & \parbox[t]{0.78\linewidth}{Nous avons roulé vers le nord pendant six heures jusqu'à un lac gelé, où}\\\emph{genre} & \parbox[t]{0.78\linewidth}{OSLO (Reuters) - A remote northern road was closed on Thursday at a frozen lake, where}\\\end{tabular}
\item[8.] \begin{tabular}[t]{@{}ll@{}}\emph{base} & \parbox[t]{0.78\linewidth}{The baker woke at four in the morning, lit the oven, and}\\\emph{detail} & \parbox[t]{0.78\linewidth}{The baker woke at five in the morning, lit the oven, and}\\\emph{para} & \parbox[t]{0.78\linewidth}{Up at four a.m., the baker got the oven going and}\\\emph{lang} & \parbox[t]{0.78\linewidth}{Le boulanger se leva à quatre heures du matin, alluma le four et}\\\emph{genre} & \parbox[t]{0.78\linewidth}{LYON (AFP) - Bakers across the region began work before dawn on Monday as flour prices}\\\end{tabular}
\item[9.] \begin{tabular}[t]{@{}ll@{}}\emph{base} & \parbox[t]{0.78\linewidth}{The children found an old map in the attic, and the next morning they}\\\emph{detail} & \parbox[t]{0.78\linewidth}{The children found an old key in the attic, and the next morning they}\\\emph{para} & \parbox[t]{0.78\linewidth}{Up in the attic the kids discovered an ancient map; the following day they}\\\emph{lang} & \parbox[t]{0.78\linewidth}{Les enfants trouvèrent une vieille carte dans le grenier, et le lendemain matin ils}\\\emph{genre} & \parbox[t]{0.78\linewidth}{DUBLIN (AP) - Children discovered a centuries-old map in a family attic, local historians said, and}\\\end{tabular}
\item[10.] \begin{tabular}[t]{@{}ll@{}}\emph{base} & \parbox[t]{0.78\linewidth}{The doctor looked at the X-ray for a long moment before she}\\\emph{detail} & \parbox[t]{0.78\linewidth}{The doctor looked at the scan for a long moment before she}\\\emph{para} & \parbox[t]{0.78\linewidth}{For a long while the doctor studied the X-ray, and then she}\\\emph{lang} & \parbox[t]{0.78\linewidth}{La médecin regarda la radiographie un long moment avant de}\\\emph{genre} & \parbox[t]{0.78\linewidth}{GENEVA (Reuters) - Hospital doctors reviewed new imaging results on Wednesday before officials}\\\end{tabular}
\item[11.] \begin{tabular}[t]{@{}ll@{}}\emph{base} & \parbox[t]{0.78\linewidth}{The storm knocked out the power, so we lit candles and}\\\emph{detail} & \parbox[t]{0.78\linewidth}{The storm knocked out the phones, so we lit candles and}\\\emph{para} & \parbox[t]{0.78\linewidth}{When the storm cut the electricity, we got out the candles and}\\\emph{lang} & \parbox[t]{0.78\linewidth}{La tempête coupa l'électricité, alors nous avons allumé des bougies et}\\\emph{genre} & \parbox[t]{0.78\linewidth}{MIAMI (AP) - A powerful storm left thousands without electricity on Saturday, and residents}\\\end{tabular}
\item[12.] \begin{tabular}[t]{@{}ll@{}}\emph{base} & \parbox[t]{0.78\linewidth}{On the first day of school, the new teacher wrote her name on the board and}\\\emph{detail} & \parbox[t]{0.78\linewidth}{On the last day of school, the new teacher wrote her name on the board and}\\\emph{para} & \parbox[t]{0.78\linewidth}{The new teacher began the first school day by writing her name on the blackboard, then}\\\emph{lang} & \parbox[t]{0.78\linewidth}{Le premier jour d'école, la nouvelle maîtresse écrivit son nom au tableau et}\\\emph{genre} & \parbox[t]{0.78\linewidth}{MADRID (Reuters) - Schools reopened on Monday, and at one primary school a new teacher}\\\end{tabular}
\item[13.] \begin{tabular}[t]{@{}ll@{}}\emph{base} & \parbox[t]{0.78\linewidth}{The mountain hut was empty except for a wooden table and a note that said}\\\emph{detail} & \parbox[t]{0.78\linewidth}{The mountain hut was empty except for a wooden chair and a note that said}\\\emph{para} & \parbox[t]{0.78\linewidth}{Apart from a wooden table and a short note, the mountain hut was deserted; the note said}\\\emph{lang} & \parbox[t]{0.78\linewidth}{Le refuge de montagne était vide, à part une table en bois et un mot qui disait}\\\emph{genre} & \parbox[t]{0.78\linewidth}{ZURICH (AP) - Rescuers who reached an empty mountain hut on Sunday found only a note that said}\\\end{tabular}
\item[14.] \begin{tabular}[t]{@{}ll@{}}\emph{base} & \parbox[t]{0.78\linewidth}{My father fixed the old radio on the kitchen table, and when it finally played,}\\\emph{detail} & \parbox[t]{0.78\linewidth}{My father fixed the old clock on the kitchen table, and when it finally played,}\\\emph{para} & \parbox[t]{0.78\linewidth}{When my father at last got the old kitchen radio working again,}\\\emph{lang} & \parbox[t]{0.78\linewidth}{Mon père répara la vieille radio sur la table de la cuisine, et quand elle joua enfin,}\\\emph{genre} & \parbox[t]{0.78\linewidth}{ROME (Reuters) - Collectors of vintage radios gathered on Saturday, and one restorer said that}\\\end{tabular}
\item[15.] \begin{tabular}[t]{@{}ll@{}}\emph{base} & \parbox[t]{0.78\linewidth}{The ship left the harbor at dawn, carrying coffee and three passengers who}\\\emph{detail} & \parbox[t]{0.78\linewidth}{The ship left the harbor at dusk, carrying coffee and three passengers who}\\\emph{para} & \parbox[t]{0.78\linewidth}{At first light the ship sailed out of port with a cargo of coffee and three travellers who}\\\emph{lang} & \parbox[t]{0.78\linewidth}{Le navire quitta le port à l'aube, chargé de café et de trois passagers qui}\\\emph{genre} & \parbox[t]{0.78\linewidth}{SANTOS (Reuters) - A cargo ship carrying coffee left port early on Friday with three passengers who}\\\end{tabular}
\item[16.] \begin{tabular}[t]{@{}ll@{}}\emph{base} & \parbox[t]{0.78\linewidth}{The library was closing, but the old man kept reading by the window until}\\\emph{detail} & \parbox[t]{0.78\linewidth}{The museum was closing, but the old man kept reading by the window until}\\\emph{para} & \parbox[t]{0.78\linewidth}{Although the library was about to close, the old man went on reading at the window until}\\\emph{lang} & \parbox[t]{0.78\linewidth}{La bibliothèque fermait, mais le vieil homme continuait à lire près de la fenêtre jusqu'à ce que}\\\emph{genre} & \parbox[t]{0.78\linewidth}{PRAGUE (AP) - The city's main library shortened its opening hours this week, and one regular reader}\\\end{tabular}
\item[17.] \begin{tabular}[t]{@{}ll@{}}\emph{base} & \parbox[t]{0.78\linewidth}{She planted tomatoes in the garden every spring, and this year}\\\emph{detail} & \parbox[t]{0.78\linewidth}{She planted roses in the garden every spring, and this year}\\\emph{para} & \parbox[t]{0.78\linewidth}{Each spring she put tomato plants in the garden; this time, though,}\\\emph{lang} & \parbox[t]{0.78\linewidth}{Elle plantait des tomates dans le jardin chaque printemps, et cette année}\\\emph{genre} & \parbox[t]{0.78\linewidth}{BORDEAUX (AFP) - Home gardeners reported a poor tomato season this year as a late frost}\\\end{tabular}
\item[18.] \begin{tabular}[t]{@{}ll@{}}\emph{base} & \parbox[t]{0.78\linewidth}{The orchestra stopped playing when the lights went out, and the conductor}\\\emph{detail} & \parbox[t]{0.78\linewidth}{The orchestra stopped playing when the alarm went off, and the conductor}\\\emph{para} & \parbox[t]{0.78\linewidth}{As soon as the lights failed, the orchestra fell silent and the conductor}\\\emph{lang} & \parbox[t]{0.78\linewidth}{L'orchestre cessa de jouer quand les lumières s'éteignirent, et le chef d'orchestre}\\\emph{genre} & \parbox[t]{0.78\linewidth}{SALZBURG (Reuters) - A power cut halted a sold-out concert on Thursday, and the conductor}\\\end{tabular}
\item[19.] \begin{tabular}[t]{@{}ll@{}}\emph{base} & \parbox[t]{0.78\linewidth}{Two hikers got lost in the forest after sunset, and one of them}\\\emph{detail} & \parbox[t]{0.78\linewidth}{Two hikers got lost in the desert after sunset, and one of them}\\\emph{para} & \parbox[t]{0.78\linewidth}{After the sun went down, two walkers lost their way among the trees, and one of them}\\\emph{lang} & \parbox[t]{0.78\linewidth}{Deux randonneurs se perdirent dans la forêt après le coucher du soleil, et l'un d'eux}\\\emph{genre} & \parbox[t]{0.78\linewidth}{DENVER (AP) - Rescue teams searched overnight for two hikers lost in a national forest, and one}\\\end{tabular}
\item[20.] \begin{tabular}[t]{@{}ll@{}}\emph{base} & \parbox[t]{0.78\linewidth}{The market was full of oranges and noise, and the woman at the corner stall}\\\emph{detail} & \parbox[t]{0.78\linewidth}{The market was full of lemons and noise, and the woman at the corner stall}\\\emph{para} & \parbox[t]{0.78\linewidth}{Oranges and shouting filled the market, and at the stall on the corner the woman}\\\emph{lang} & \parbox[t]{0.78\linewidth}{Le marché était plein d'oranges et de bruit, et la femme à l'étal du coin}\\\emph{genre} & \parbox[t]{0.78\linewidth}{SEVILLE (Reuters) - Orange prices rose sharply at local markets this week, and one stallholder}\\\end{tabular}
\end{itemize}}

\section{Toolkit}
The code runs on a single laptop and measures, for any Hugging Face causal language model: the prompt offset under shared continuations and its signal, noise and mean-squared versions, with the unbiased estimator, centre-labelled octave windows, global and local exponents, and bootstrap intervals over prompt pairs (\texttt{real\_run.py}; \texttt{scale\_run.py} for the hand-written scenes); the non-degeneracy check of Assumption~\ref{ass:nondeg} (\texttt{prenorm\_check.py}); the exact direct/relay decomposition of the attention writes together with uniform and per-head dilution compensation (\texttt{real\_mech.py}); the trade-off between retention and perplexity with a placebo control (\texttt{tradeoff\_real.py}); and the single- and multi-turn instruction-adherence tests (\texttt{behavior\_test.py}, \texttt{multiturn\_test.py}). The toolkit, together with the exact real-corpus prompt pairs, will be released publicly.

\end{document}